\documentclass[10pt,a4paper]{article}

\usepackage[english]{babel}

\usepackage[top=2.5cm,bottom=2.5cm,left=3cm,right=3cm,marginparwidth=2.5cm]{geometry}
\usepackage{amsmath, amssymb, amsthm}
\usepackage{physics}

\DeclareMathOperator*{\argmax}{\arg\!\max} % declare argmax
\NewDocumentEnvironment{alignb}{b}{%
  \begin{align*}
  \refstepcounter{equation} #1 \tag{\theequation}
  \end{align*}
}{} % allows align* environment with a tag at the last line of the equation.
\usepackage{bm} % Bold greek letters
\usepackage{dsfont} % Enables to type \mathbb-style letters
\usepackage{cite}
\usepackage[linesnumbered,noend,lined,boxed,commentsnumbered,ruled,longend]{algorithm2e} % Allows including algorithms
\usepackage{multicol}
\usepackage{graphicx}
\graphicspath{{./Fig/}}
\usepackage[dvipsnames]{xcolor} % Access to colors with preset names
\usepackage{soul} % for highlighted texts
\sethlcolor{Yellow}
\usepackage{subcaption}
\DeclareCaptionFont{small}{\small}
\usepackage{hyperref}
\hypersetup{
    colorlinks=true,
    linkcolor=NavyBlue,
    filecolor=OrangeRed,      
    urlcolor=Emerald,
    citecolor=OrangeRed,
    pdftitle={Draco-Eunjeong},
    }
\usepackage{charter} %charter, mathpazo, fourier, noto
\usepackage[T1]{fontenc}
\usepackage[utf8]{inputenc}
\usepackage[toc]{appendix}
\usepackage{apptools} % For Lemma numbering in appendices
\usepackage{centernot} % used to negate some symbols to clarify the "don't care" variables
\AtAppendix{\counterwithin{lemma}{section}}

\newtheorem{assumption}{Assumption}

\newtheorem{lemma}{Lemma}
\newtheorem{proposition}{Proposition}
\newtheorem{theorem}{Theorem}
\newtheorem{definition}{Definition}

\SetKwInput{KwData}{Input}
\SetKwInput{KwResult}{Output}
\SetKwProg{Init}{Initialize}{}{}
\SetKwFor{ParFor}{for}{do in parallel}{end for}
\SetKwRepeat{Do}{do}{while}
\SetCommentSty{mycommfont}

\newcommand{\cache}[1]{} % make multiple lines into comments

\title{DRACO: Decentralized Asynchronous Federated Learning over Row-Stochastic Wireless Networks\footnote{This paper has been submitted to a peer-reviewed journal and is currently under review.}}
\author{Eunjeong Jeong$^1$ \and Marios Kountouris$^{2,3}$}
\date{\small{
    $^1$Department of Computer and Information Science, Linköping University, Sweden\\
    $^2$Andalusian Research Institute in Data Science and Computational Intelligence (DaSCI),\\ Department of Computer Science and Artificial Intelligence, University of Granada, Spain\\
    $^3$Communication Systems Department, EURECOM, 06904 Sophia Antipolis, France\\%[2ex]%\today
}}

\begin{document}
\maketitle

%\tableofcontents%
%\pagebreak

\begin{abstract}
    Emerging technologies and use cases, such as smart Internet of Things (IoT), Internet of Agents, and Edge AI, have generated significant interest in training neural networks over fully decentralized, serverless networks. A major obstacle in this context is ensuring stable convergence without imposing stringent assumptions, such as identical data distributions across devices or synchronized updates.
    In this paper, we introduce DRACO, a novel framework for decentralized asynchronous Stochastic Gradient Descent (SGD) over row-stochastic gossip wireless networks. Our approach leverages continuous communication, allowing edge devices to perform local training and exchange model updates along a continuous timeline, thereby eliminating the need for synchronized timing. Additionally, our algorithm decouples communication and computation schedules, enabling complete autonomy for all users while effectively addressing straggler issues.
    Through a thorough convergence analysis, we show that even without predefined scheduling policies, DRACO achieves high performance in decentralized optimization while maintaining low variance across users. Numerical experiments further validate the effectiveness of our approach, demonstrating that controlling the maximum number of received messages per client significantly reduces redundant communication costs while maintaining robust learning performance. 
\end{abstract}

\section{Introduction}\label{sect:intro}

Recent advancements in machine learning, networked intelligent systems, and wireless connectivity have paved the way for various innovative applications and use cases across various sectors, including the Internet of Things (IoT), consumer robotics, autonomous transportation, and edge computing. These systems increasingly rely on decentralized learning architectures for processing data where generated, minimizing latency and bandwidth usage while enhancing privacy. However, these benefits come with significant challenges, particularly in terms of ensuring efficient and reliable communication and processing within inherently unstable and diverse network environments. Addressing these challenges requires novel approaches that adapt to the unique demands of decentralized architectures, fostering robust and expandable solutions for real-time data processing and learning.
In this work, we consider the problem of communication efficiency in federated learning (FL) \cite{mcmahan17:FL} and in particular in serverless (fully decentralized) learning settings that operate without a central coordinating server \cite{nedic09, lalitha18-FullDecentFL, roy19, karras22, qin22}. Asynchronous learning, empowering each participant to conduct local training and data transmission at their own pace, is a standard and relevant design choice in decentralized network schemes \cite{nadiradze21, dai22-dispfl, esfandiari21, liang20, kanamori23}. Asynchronous and decentralized learning have an advantage when used separately from each other, manifesting as adaptability to limited resources and downsized communication overhead. Yet unfortunately, when these two paradigms are combined, their integration poses a greater challenge in achieving a unanimous global consensus, as required for instance in the development of sophisticated navigation algorithms \cite{beltran23-survey}.

Decentralized optimization studies in the literature often involve high ``synchronization costs'' due to the complexity of ensuring consensus. In other words, the majority of asynchronous learning schemes are executable only if all participants have a common sense of the global communication rounds, which have to be, in a way, synchronously counted. This paradoxical agreement in synchronized clocks takes an additional cost to bear while carrying out related techniques over wireless networks with message losses or delays due to unstable channel conditions.
As a result, the focus has shifted towards analyzing decentralized learning using asynchronous gossip protocols \cite{zhang21, hegedus21, jeong22-async, wulfert23, even24}. The introduction of gossiping, leveraging random or probabilistic communication, has rendered the proposed algorithms more compelling than those relying on predefined schedules, mainly thanks to their resilience to dynamically changing connectivity \cite{blot16}.

However, existing studies on asynchronous gossip optimization usually adopt several strong assumptions that make the proposed solutions less applicable in realistic scenarios, in particular when involving wireless communication protocols. Early works on distributed optimization have relied on doubly stochastic weights, which are suitable only for undirected or balanced networks \cite{nguyen23}. Despite the prevalence of the assumptions among many related studies, algorithms designed with doubly stochastic weights are unable to be constructed over arbitrarily directed graphs \cite{xin19:frost}.
Distributed optimization over directed graphs has been extensively studied in control theory \cite{nedic15, akbari17, li18, ghaderyan24}. More recently, federated learning (FL) research has started addressing challenges related to asymmetric connectivity, using row-stochastic matrices \cite{mai16:rowst-distopt} and time-varying directed graphs \cite{nguyen24:decentfl-directed}. Some works have even incorporated personalization \cite{liu24:directedPFL}. However, a significant research gap remains in exploring how collaborative learning networks can maintain robustness in the presence of unreliable transmissions.

%{In practice, decentralized learning over networks with row-stochastic gossip communication patterns is a separate research topic branched out from those studies over doubly stochastic graphs.}

\begin{figure}[t]
    \centering
    \includegraphics[width=\textwidth]{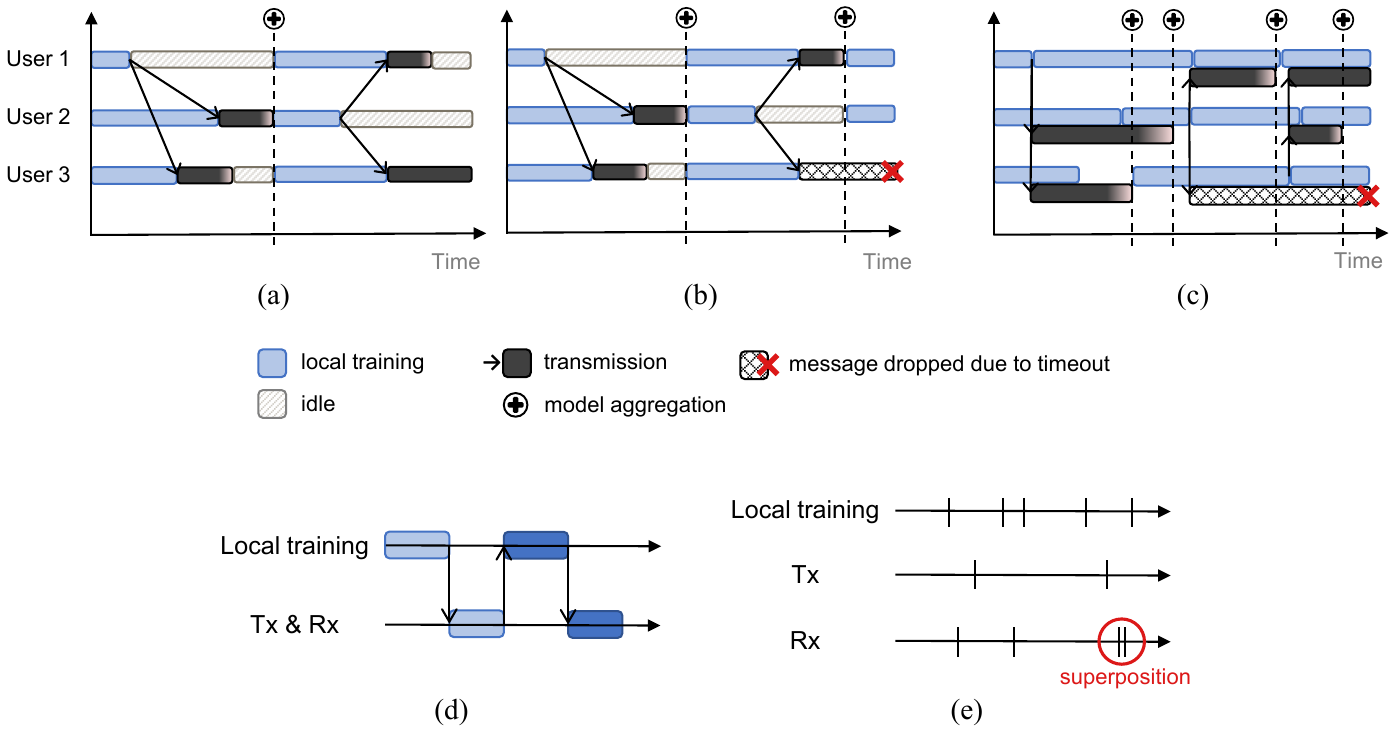}
    \caption{A schematic view of DRACO's timelines with comparisons. (a) Synchronous FL; (b) asynchronous FL with transmission delay deadline; (c) (in DRACO) fully asynchronous FL with delay deadline, but the iteration count is continuous;     
    (d) sequential computation and communication over a doubly stochastic network; (e) timelines of DRACO with decoupled computation and communication over a row-stochastic network. If two messages arrive at the same agent with a negligibly small time gap (in red circle), they are considered simultaneous and are used for the same model aggregation step. The concept of superposition window is elaborated in Section \ref{subsect:commsys}.
    }
    \label{fig:overview}
\end{figure}
These stringent assumptions have been introduced intentionally or inevitably since decentralized learning over gossip communication presents several technical challenges. One of the major challenges is uncertainty in convergence; for instance, irregular or non-uniform communication probabilities in the gossip network can lead to variable convergence rates or convergence to suboptimal solutions.
Furthermore, the performance of decentralized learning algorithms, for instance in terms of convergence rate and communication load per iteration, is more sensitive to the specific network topology or graph \cite{song22}. Small changes in communication probabilities or network configuration could significantly impact learning dynamics. Therefore, addressing these difficulties requires specialized algorithmic solutions and in-depth analyses to guarantee the effectiveness, stability, and convergence of asynchronous learning over decentralized networks.
This involves developing algorithms that adapt to irregular communication probabilities while maintaining robustness and efficiency across diverse network structures.

In this work, we introduce DRACO, a novel framework for decentralized asynchronous FL. In brief, we propose a scheme characterized by two foundational elements: (i) it facilitates continuous, asynchronous operations without a global iteration count, and (ii) it employs decoupling communication and computation strategies, integrated with gradient pushing.

Firstly, our asynchronous learning approach operates continuously, permitting messages to be sent and received at non-uniform, non-integer time instants. This flexibility translates into that message arrivals are not confined to specific multiples of a global round duration, allowing each node to operate independently, based on its own schedule. The variability in each client's timeline is illustrated in Fig. \ref{fig:overview}c. For instance, while User 3 is engaged in its second local updating round, User 1 is already progressing through its third round. Although this approach makes it difficult to trace the progress of each client's model at any given moment due to the variability in their timelines, it significantly shortens their idle time, enhancing the efficiency of the learning process. To alleviate the impact of outdated gradients that may impede local models from being optimized, messages that exceed a certain delay threshold are disregarded.

Secondly, DRACO leverages decoupled communication and computation schedules, as illustrated in Fig. \ref{fig:overview}e. In a fully asynchronous network, the integrated learning process is less likely to stagnate when local training and transmission occur independently. In an environment where all users are busy, as in Fig. \ref{fig:overview}c, if they always forward their new reference models after aggregating local updates from its neighbors (one-hop senders) like in Fig. \ref{fig:overview}d, that way of exchange jeopardizes the optimization by communication overloads twice as heavy as push-based collaboration. Furthermore, the content delivered to each other can often be duplicated or overwritten. Thus, separating the two types of schedules departs from conventional methodologies that mandate a sequential or predetermined order for gradient updates and gossip communications.

This paper introduces an asynchronous learning framework within a fully decentralized network, accommodating asymmetric communication weight graphs. Our approach distinguishes itself from existing works in several key aspects. First, it introduces an asynchronous and decentralized learning model in a continuously defined timeline, removing the need for quantized transmission schedules. Consequently, our proposed technique exhibits adaptability to dynamic network conditions.

Second, we introduce a novel and more realistic approach to addressing asynchrony in intelligent wireless networks. Rather than limiting the analysis to comparisons between synchronous and asynchronous communication or centralized and decentralized learning, our study embraces asynchrony and the absence of a central server as inherent challenges. In this context, we aim to investigate whether our proposed framework can substantially improve user performance.
%{Second, our scheme simplifies participants' actions, providing a straightforward and easy-to-implement framework for each user. This simplicity not only streamlines the learning process but also contributes to the scalability of our approach.}
Additionally, we address the uncertainty and variability inherent in wireless networks, enhancing the scheme's resilience to fluctuations in connectivity. By incorporating these features, our work contributes a unique perspective to decentralized learning, offering a practical and efficient solution for real-world scenarios.

\subsection{Related Works}\label{sect:related_works}

\textbf{Asynchronous decentralized learning}\quad
In synchronous learning systems, all participants ought to wait for the slowest learner, known as a straggler, before proceeding to the next global round. As depicted in Fig. \ref{fig:overview}b, asynchronous learning with a transmission delay deadline effectively reduces the overall training time of synchronous systems by excluding users whose updates arrive after a predetermined deadline \cite{jeong22-async, xing21}. This approach is applied not only to asynchronous settings but to synchronous learning through partial participation \cite{wang20-neurips}. Both asynchronous learning and partially participating synchronous learning face the challenge of variance reduction since only a subset of local updates is considered in each training round \cite{jhunjhunwala22, li23, qin23, cho23}. Despite fewer average participation in model aggregation per user compared to synchronous methods, asynchronous learning performs as well as its counterpart, especially in solving large-scale multi-user optimization problems \cite{even22}. Nevertheless, this approach requires users to start their computations simultaneously to synchronize the global phase, leading to idle times when a message arrives before the start of the next iteration. Additionally, sufficient local storage is necessary to manage multiple messages queued in the receive buffer until the next round.

\textbf{Randomized communication over serverless and directed networks}\quad
Recent studies in decentralized learning have explored algorithms implementable for networks modeled by directed graphs, where the connectivity matrix is not necessarily doubly stochastic. This adaptation is often necessary when neither full-duplex nor half-duplex systems can ensure stable gradient transmissions. Techniques, such as push-sum \cite{rabbat14, tsianos12:pushsum, taheri20, qureshi22:push-saga, toghani23-parspush, assran19:gradientpush, assran21}, push-pull \cite{nedic16, hsieh23}, and random walk \cite{mao20:walkman, ayache19:randomwalk-dsgd, hendrikx23}, have been proposed to improve decentralized optimization on directed graphs.
Meanwhile, row-stochastic communication \cite{nedic20} significantly reduces both the number of communication rounds and storage requirements on edge devices; hence, this benefits in tackling complex problems, specifically those involving small-scale neural networks \cite{ghaderyan24}. 
Among random communication protocols, gossip protocol is well-known for its rapid information spread but also criticized for its high network resource consumption \cite{giaretta19}. Consequently, asynchronous gossip learning in such contexts needs innovative approaches to manage information flow among edge devices \cite{gholami24:digest}.

\textbf{Decoupling communication and computation}\quad
Unlike traditional methods that align gradient computation and communication either sequentially or in parallel, decoupling these processes significantly accelerates peer-to-peer averaging by releasing clients from waiting for others \cite{nabli23-dadao, nabli23:a2cid2, belilovsky20}. In AD-OGP~\cite{jiang21}, the authors replaced global communication slots with an event-based aggregation system, encompassing activities such as prediction and local updating.
This unified timeline of events is particularly well-suited for environments where users train locally at different computational speeds. However, the event types of AD-OGP are restricted to ``prediction'' and ``local updating'', overlooking the impact of transmission delay. The authors assume that message delay, defined as the time gap between the latest prediction and the local updating event within a user, provides no insight into how long it takes for a message to reach a neighboring node. Despite the growing interest in approaches for effective timeline integration, only a few studies have explored decoupled model averaging over unreliable wireless networks, where issues such as packet loss or delays are prevalent.

%=============================================================
\section{System Model}\label{sect:system_model}
We consider the following optimization task over $N$ clients whose goal is to minimize
\begin{align}
    f(\mathbf{x}) := \frac{1}{N}\sum_{i\in\mathcal{U}} f_i(\mathbf{x})
    \label{fn:obj}
\end{align}
where $\mathbf{x}\in \mathbb{R}^d$ is an $d$-dimensional model parameter and $\mathcal{U}=
\{1,\cdots,N\}$ is the set of network users. In a serverless network, there is no global model $\mathbf{x}_t$; instead, each agent $i$ holds $\mathbf{x}_t^{(i)}$, which serves as a reference for the globally acquired model. Therefore, the objective function can be rewritten as
\begin{align}
    \mathbf{x}^*=\inf_{\mathbf{x}\in\mathbb{R}^d} \sum_{i=1}^N f_i(\mathbf{x}^{(i)})\ .\label{fn:obj-2}
\end{align}

To tackle the minimization problem described in (\ref{fn:obj}) or (\ref{fn:obj-2}), we adopt a decentralized stochastic gradient descent (DSGD) approach. In this approach, individual devices iteratively enhance their local models $\mathbf{x}^{(i)}$ and subsequently share these estimates with their neighboring nodes, which in turn could vary over time.

\subsection{Absence of a global belief}
The underlying assumption regarding the global consensus is that each user cannot reach a global ``true parameter'', denoted by $\mathbf{x}^\ast$, by local updates only. The global model $\mathbf{x}$ should be a vector combined with the beliefs (pseudo-global model) at each agent, said $\mathbf{x} = \{\mathbf{x}^{(1)}, \mathbf{x}^{(2)}, \cdots, \mathbf{x}^{(N)}\}$. However, in practice, none of the agents works as a central server or aggregator, which can obtain a centralized global model. We therefore adopt a virtual global model $\mathbf{\bar{x}}$ that could have been acquired through the superposition of all beliefs if the network had an entirely authorized server, i.e.,
\begin{align*}
    \mathbf{\bar{x}} = \mathbb{E}_{i\in \mathcal{U}}[\mathbf{x}^{(i)}] = \frac{1}{N}\sum_{i=1}^N \mathbf{x}^{(i)}.
\end{align*}
Therefore, when $P$ seconds have elapsed since the initial moment $t_0$, the virtual global model difference between these two time instances can be expressed as
% expectation of model update during $P$
\begin{align*}
    \mathbf{\bar{x}}_{t_0+P}-\mathbf{\bar{x}}_{t_0} = \frac{1}{N}\sum_{i=1}^N \big( \mathbf{x}_{t_0+P}^{(i)}-\mathbf{x}_{t_0}^{(i)} \big).
\end{align*}

\subsection{Communication system}\label{subsect:commsys}
In our work, the processes of computation and communication are decoupled, hence when to locally train and when to transmit the updates are determined independently at each user.
Since there can be infinite instants between any two close events on the continuous timeline, each message is likely to arrive at a different moment. Thus, practically speaking, there is no aggregation during the entire process even though two updates arrive at the same destination node by a narrow margin of time. In this regard, we introduce a \emph{superposition window}, which is analogous to congestion windows in TCP (Transmission Control Protocol) \cite{postel81:tcp}. Similar to a TCP window, the superposition window in DRACO controls the flow of received updates by grouping the messages for one aggregation. This leads to lower computation costs due to the fact that renewal of the local reference model every time a message arrives is avoided.

%In this paper, we investigate the effect of (i) topology and (ii) wireless channels on DRACO. The first case, which involves a time-invariant connectivity graph, covers scenarios in which the topology is fixed throughout the learning process. For any two user nodes connected to an edge, a message sent from one node is always successfully received at the other. The physical distance (i.e., geographical coordinates) is not considered. Under this setting, we study the impact of the frequency of successfully received messages and the characteristics of the connectivity graphs. %On the other hand, in the second case, the connectivity graph changes over time. %In this setup, user nodes are randomly positioned, with their coordinates following a uniform distribution. Information exchange between nodes is affected by factors such as interference and channel capacity constraints. Differentiating from the fixed topology scenarios, we mainly verify the influence of the transmission duration deadline, superposition window, and unification period (further discussed in Section \ref{subsect:uni}) accounting in general for all wireless channel conditions.  

This paper investigates the effect of unreliable wireless communications and controlled transmissions on the performance of DRACO. Specifically, we consider scenarios with time-invariant connectivity graphs, which represent stable network topologies during the learning process. This simplified assumption facilitates a focused analysis of how the frequency of successful message receptions and the underlying structure of the communication network affect convergence in the learning process. 
Unlike traditional fixed-topology models, we explicitly account for the inherent unreliability of wireless channels. In our model, successful message delivery between connected nodes is probabilistic and not guaranteed, influenced by factors such as physical distance, interference, and channel capacity limitations. User nodes are randomly distributed and their geographical positions directly affect communication probabilities. 
To provide a more comprehensive understanding beyond standard fixed-topology analyses, we examine the effect of the frequency of successful message receptions within a defined unification period (detailed in Section~\ref{subsect:uni}). This approach provides a more nuanced assessment of the learning process under various wireless channel conditions.

A weighted graph at a certain instance %{at the moment of an event $k$}
is mathematically defined as a $N\times N$-sized matrix where each element indicates whether $i$ transmits its message to one of its neighbors $j$ or not. It follows a conditional probability distribution if there is a communication event on client $i$. Transmission incidents are defined as
\begin{align}
    q_k^{i} &= \begin{cases}
    1, & \text{if $i$ broadcasts $\Delta^{(i)}$ at $k$}\\
    0, & \text{otherwise}.
    \end{cases}\\
    q_k^{ij} &= \begin{cases}
    1, & \text{if $j$ receives $i$'s message sent at $k$}\\
    0, & \text{otherwise},
    \end{cases}
\end{align}
where $k$ is the index of an event. We define the neighborhood of user $i$, denoted by $\mathcal{N}_t(i) = \{j| q_t^{ij}=1\}$, as the set of all users $j$ that have an edge going from $i$ to $j$ at time $t$. It is also possible to denote the neighbor set with respect to event $k$, such as $\mathcal{N}_k(i)$.
Following the notation in \cite{wang20-neurips}, these participation indicators are normalized across all moments, i.e., $\sum_{j\in\mathcal{U}\setminus\{i\}} q_t^{ij} = 1$ for $q_t^{ij}\geq 0$ and for all $i,\ t$. In addition to the definition, we define $\rho<1$ that satisfies $\sum_{j\in\mathcal{U}\setminus\{i\}} (q_t^{ij})^2 \leq \rho^2$ for all $i,\ t$.

%{A broadcasting event is a necessary condition for a reception event. Let $e$ indicate the event of a client node $i$ broadcasting its local update at time $t$. When this event $e$ occurs, the other nodes receive the message based on a conditional probability distribution, which also implies that a set of neighbors (recipients) is decided after the occurrence of $e$. If by $\delta_i = 1$ we indicate the incidence that node $i$ has broadcast a message at a given instant, the probability that another node $j$ receives this message is conditioned on this event $e$. In other words, $P\left[\delta_{ij}=1 | \delta_i=0\right] = 0$ for all $i,j$.}

\subsection{Local gradient computations}
Each user performs stochastic gradient computations by iterating $B$ batches of the local training datasets. $\Delta$ represents the local update of the model, defined as the difference between the model's state prior to the mini-batch training and its state after completing training on $B$ batches of training samples.
%The difference between the $B$-th calculation and the current model, denoted by $\Delta$, is saved as the latest version of the local update.

\begin{assumption}\label{assmp:time_comp}{(Exponential local gradient computation time.)}
    \normalfont The computation time $\tau_i$ of the stochastic gradient $g_i(\mathbf{x})\in\mathbb{R}^d$ at user $i$ is exponentially distributed, i.e., $\tau_i\sim exp(\lambda_i)$.
\end{assumption}

In the context of point processes, one can consider a PPP along the real line by examining the count of points within a specific interval $(t_0, t_0+P]$ \cite{kingman92:poisson}. For a homogeneous PPP with rate parameter $\lambda>0$, the likelihood that the count of points, denoted by $num(t_0, t_0+P]$, equals a certain integer $m$ can be described by the following expression:
\[
\Pr\{num(t_0, t_0+P]=m\} = \frac{(\lambda P)^m}{m!}e^{-\lambda P}.
\]
This formula calculates the probability of exactly $m$ occurrences within the interval based on the Poisson distribution, where $\lambda P$ represents the expected number of points in the interval and $e^{-\lambda P}$ adjusts for the total rate of occurrences over the span.

\section{DRACO: Proposed Decentralized Asynchronous Learning}\label{sect:draco}

The main rationale behind the proposed algorithm is to provide an answer to the following question: \emph{How can we issue instructions to each user in the absence of a global time loop in the network?} To resolve this issue, we design the system such that each node focuses solely on its actions without considering the training progress or the channel conditions of the other nodes. Defining the algorithm within a unified time loop in asynchronous and fully decentralized networks presents several practical limitations in real-world systems. 
A significant challenge lies in the absence of a consistent global time reference, such as global iteration rounds, denoted as $t$, or timestamps marking the completion of each user's local computations, marked as $k$. This inconsistency arises because the total number of local training iterations varies across users, even when their updates are observed simultaneously. As a result, if the algorithm mandates exchanges every $t_P$ seconds or every $k_P$ global slots, some local models may fall behind in development due to completing fewer local training steps compared to others.
To address this issue, we avoid defining the procedure as either sequential or simultaneous. Our algorithm adopts instead a unified global loop where all users work in parallel. This global loop effectively encapsulates the learning process conducted by each user.

\begin{figure*}[t]
    \centering
    \includegraphics[width=0.7\textwidth]{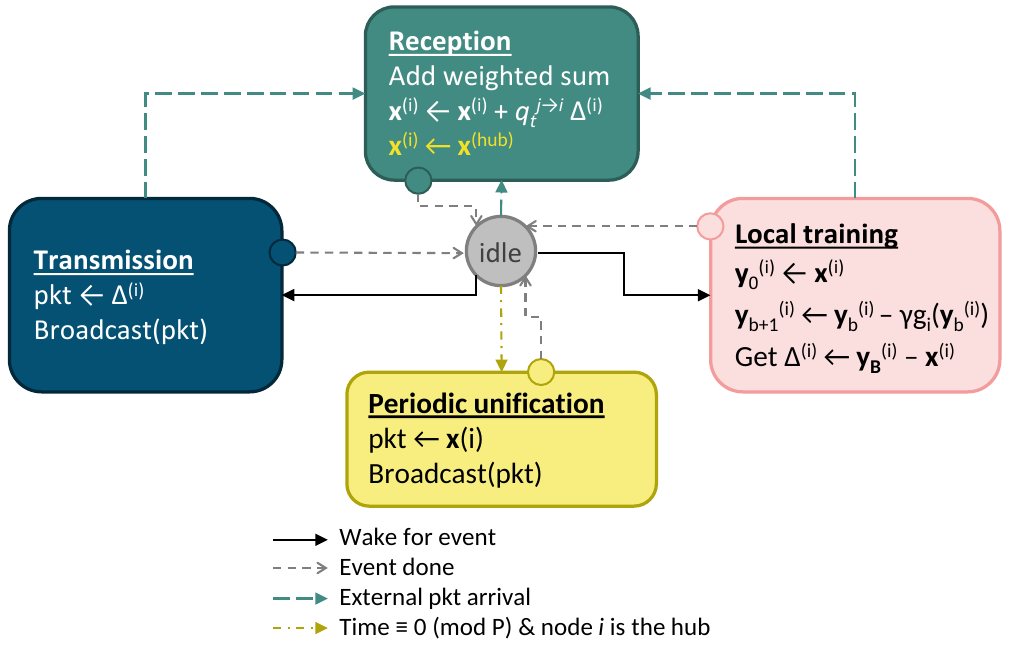}
    \caption{The proposed algorithm (DRACO) in a chain graph illustrating the states of possible actions within each agent $i$.}
    \label{fig:chaingraph}
\end{figure*}

\begin{figure}[t]
  \centering
  \scalebox{0.95}{
    \begin{minipage}{0.95\linewidth}
      \begin{algorithm}[H]
        \SetAlgoLined
        \caption{\label{alg:draco-userview} User-centric algorithm of DRACO. A pseudo-algorithm for source code reproduction is provided in Appendix \ref{appx:psuedo-alg}.}
    \BlankLine
    \DontPrintSemicolon
    \KwData{$\gamma, \mathbf{x}_0, B, T$, $P$}
    \KwResult{$\{\mathbf{x}_t : \forall t\}$}
    \ParFor{$i=1,\cdots,N$}{
    \While{$t<T$}{
        $t\leftarrow\text{clock}()$\;
        \If{there is an event at time $t$}{
            \uIf{grad computation step}{
                $\mathbf{y}_{0}^{(i)}\leftarrow \mathbf{x}^{(i)}$\;
                \For{$b=0,\cdots,B-1$}{
                    $\mathbf{y}_{b+1}^{(i)}\leftarrow \mathbf{y}_b^{(i)}-\gamma g_i(\mathbf{y}_b^{(i)})$\;
                }
                $\Delta^{(i)}\leftarrow \mathbf{y}_B^{(i)} - \mathbf{x}^{(i)}$ \tcp*{local batch training}
            }
            \ElseIf{transmission step}{
                $i$ sends $\Delta^{(i)}$ to its neighbors \;
                \For{$j\in\mathcal{N}(i)$}{
                    $j$ receives $\tilde{\Delta}^{(i)}$\;
                    $\mathbf{x}^{(j)} \leftarrow \mathbf{x}^{(j)} + \sum_{i\in\mathcal{U}} q_t^{i\rightarrow j}\tilde{\Delta}^{(i)}$ \tcp*{aggregation}
                }
                }
            }
        \If{$t\equiv0$ (mod $P$) and $t>0$ and $i$ is the hub at $t$}{
            $i$ broadcasts $\mathbf{x}^{(i)}$\;
            \For{$j\in\mathcal{U}\setminus\{i\}$}{
                $j$ receives $\tilde{\mathbf{x}}^{(i)}$\;
                $\mathbf{x}^{(j)}\leftarrow \tilde{\mathbf{x}}^{(i)}$ \tcp*{unification}
            }
            %\hl{$\mathbf{x}^{(i)}\leftarrow\mathbf{x}^{(i)}+(\gamma-1)\mathbf{u}^{(i)}$} \tcp*{amplify}
            %\hl{$\mathbf{u}^{(i)}\leftarrow \mathbf{0}$}\;
            }
        }
    } \Return $\left(\mathbf{x}_T^{(i)}\right)_{1\leq i\leq N}$\;
    \end{algorithm}
    \end{minipage}
    }
\end{figure}

At each instance, every user selects one of the following three statuses based on a probability distribution: (1) remaining idle, (2) transmitting a message to neighboring nodes, or (3) conducting local model training. Local computation involves batch training iterated $B$ times to compute the update, termed $\Delta$. During transmission, the user broadcasts its local update. If a node recognizes delivery from the other nodes, it switches to a fourth (4) status (receiving mode), renewing its reference model by aggregating the model updates from neighboring nodes. Unlike these four statuses, a node turns to the fifth (5) status when a periodic timeout occurs. As depicted in the yellow box in Figure \ref{fig:chaingraph}, a temporary hub broadcasts its reference model instead of a local update when the time is a multiple of the period $P$. The corresponding explanation as a form of algorithm is provided in Algorithm \ref{alg:draco-userview} on page \pageref{alg:draco-userview}.

Note that the `idle' state is included since we assume that the agents alter their states instantaneously, i.e., without delay. The node's status is considered idle when it does none of the aforementioned steps. However, in practice, any activity takes time to complete, implying that each timestamp represents the moment that each action just finished. By this interpretation, the participants do not have an actual break time in practical scenarios, which is also applied to the experiments in Section \ref{sect:exp}.

\textbf{Notations.}\quad
$\mathcal{U}$ represents the set of participants within the network with $\mathcal{Q}:=\{q_t^{ij}\}$ for all $i,j\in\mathcal{U}$ and $t$. Also, $\sum_{j\neq i}$ represents the summations of variables attributed to any user other than user $i$, i.e., $j\in\mathcal{U}\setminus\{i\}$. A user $i$'s local update at time $t$ is symbolized as $\Delta_t^{(i)}$. When a user $i$ sends $\Delta^{(i)}$ or $\mathbf{x}^{(i)}$, the recipient $j$ receives $\tilde{\Delta}^{(i)}$ or $\tilde{\mathbf{x}}^{(i)}$, which are identical to the sender's original contents if the transmission is free from distortion. Throughout this manuscript, the term `update' is used only as a noun that signifies the result derived from the difference between a local reference model and a newly obtained model through batch training. To avoid potential confusion, any instances in this paper that involve the action of updating are called alternatively, such as `renew' or `iterate on'.

\subsection{Periodic Unification}\label{subsect:uni}
Local models are likely to diverge when the network does not use a central server because no one synchronizes its different learning stages. Like conventional FedAvg, periodic unification can effectively resolve the variance-reduction problem among local reference models.
A countable upper bound for the number of messages per unit time is required for analysis because otherwise, the losses diverge to infinite. It is also reasonable to assume that it is finite because, in real-life applications, messages are countable even though the number of definable instances is infinite. Based on this, Assumption \ref{assmp:finite_Psi} and Definition \ref{def:psi} are introduced as follows.

\begin{assumption}{(Finite number of messages during a unit time period)}\label{assmp:finite_Psi}
    During every period $P$, the number of messages that each user receives is finite.
\end{assumption}

\begin{definition}{(Maximum number of receiving messages per user)}\label{def:psi}
    Let $\psi_i(t_\text{start}, t_\text{end})$ indicate the function that counts the number of messages arrived at user $i$ since time $t_\text{start}$ until time $t_\text{end}$. For any $i\in\mathcal{U}$ and $m\in[0,1,\cdots, \lfloor \frac{T}{P}\rfloor-1]$ , 
    \[
    \psi_i(mP, (m+1)P)\leq \Psi\ ,
    \]
    where $\Psi$ is the maximum number of messages that a user permits to receive during time duration $[mP, (m+1)P)$.
\end{definition}
The $\Psi$ term not only justifies the number of messages to be countable but also functions as a communication budget per period. Interestingly, when a decentralized network has a fixed communication budget per unit time, performing many consensus steps can effectively reduce the error even though each gossiping step renders low precision. \cite{hashemi20}

%==================== ANALYSIS =============================
\section{Convergence Analysis}\label{sect:analysis}
In this section, we analyze the convergence performance of DRACO. For that, following the common practice in the literature, we make the subsequent assumptions along with the objective function.

\begin{assumption}{(Lipschitz gradient.)}\label{assmp:lipschitz} For any $\mathbf{x}, \mathbf{y}\in\mathbb{R}^d$ and for any $i\in\mathcal{U}$, there is a nonnegative $L$ that satisfies
    \begin{align}
        \| \nabla f_i(\mathbf{x}) - \nabla f_i(\mathbf{y}) \| \leq L\|\mathbf{x}-\mathbf{y}\|.%,\ \forall \mathbf{x},\mathbf{y},\ i. 
    \end{align}
\end{assumption}

\begin{assumption}{(Unbiased stochastic gradient with bounded variance.)} \label{assmp:sg} For all $\mathbf{x},\ i$,
    \begin{align}
        \mathbb{E}[g_i(\mathbf{x}) | \mathbf{x}] = \nabla f_i(\mathbf{x}) \ \text{and}\ 
        \mathbb{E}\left[ \| g_i(\mathbf{x}) - \nabla f_i(\mathbf{x}) \|^2 | \mathbf{x} \right] \leq \sigma^2
    \end{align}
\end{assumption}

\begin{assumption}{(Bounded gradient divergence.)}\label{assmp:zeta} For all $t\in[0, T)$ and $i\in\mathcal{U}$, the gradient divergence is bounded by $\zeta$, i.e.,
    \begin{gather}
        \| \nabla f_i(\mathbf{x}_t^{(i)})-\nabla f(\mathbf{x}_t) \|^2 \leq \zeta^2 .
    \end{gather}
\end{assumption}

From Assumption \ref{assmp:zeta}, an alternative deviation of local gradients is derived as in Lemma \ref{lemma:dev_local_grads}.
\begin{lemma}{(Deviation of local gradients)} When $N>4$, for all $\mathbf{x},\ t$, \label{lemma:dev_local_grads}
    \begin{align*}
        \Big\| \sum_{j\in\mathcal{U}} q_t^{j\rightarrow i} \big[\nabla f_i(\mathbf{x}_t^{(i)}) - \nabla f_j(\mathbf{x}_t^{(j)})\big] \Big\|^2 \leq \frac{2N\zeta^2}{N-4}\ .
    \end{align*}
\end{lemma}

\begin{proof}
The left side of the inequality above can be rephrased as
\begin{align*}
     \Big\| \sum_{j\in\mathcal{U}} q_t^{j\rightarrow i} \big[\nabla f_i(\mathbf{x}_t^{(i)}) - \nabla f_j(\mathbf{x}_t^{(j)})\big] \Big\|^2=  \Big\| \nabla f_i(\mathbf{x}_t^{(i)}) - \sum_{j\in\mathcal{U}} q_t^{j\rightarrow i}\nabla f_j(\mathbf{x}_t^{(j)})\Big\|^2. %\\&= \sum_{j\in\mathcal{U}} (q_t^{j\rightarrow i})^2 \big\| \nabla f_i(\mathbf{x}_t^{(i)}) - \nabla f_j(\mathbf{x}_t^{(j)}) \big\|^2 .
\end{align*}
By adding and subtracting $\nabla f(\mathbf{x}_t)$, we have
\allowdisplaybreaks{\begin{align*}
    &\Big\| \nabla f_i(\mathbf{x}_t^{(i)}) - \sum_{j\in\mathcal{U}} q_t^{j\rightarrow i}\nabla f_j(\mathbf{x}_t^{(j)})\Big\|^2 \\
    &=\Big\| \nabla f_i(\mathbf{x}_t^{(i)}) -\nabla f(\mathbf{x}_t) +\nabla f(\mathbf{x}_t) -\sum_{j=1}^N q_t^{j\rightarrow i}\nabla f_j(\mathbf{x}_t^{(j)}) \Big\|^2\\
    &=\Big\| \nabla f_i(\mathbf{x}_t^{(i)}) -\nabla f(\mathbf{x}_t) +\frac{1}{N}\sum_{i'=1}^N\nabla f_{i'}(\mathbf{x}_t^{(i')}) -\frac{1}{N}\sum_{i'=1}^N\sum_{j=1}^N q_t^{j\rightarrow i}\nabla f_j(\mathbf{x}_t^{(j)}) \Big\|^2\\
    % (a+b)^2 \leq 2a^2 + 2b^2
    &\overset{(a)}{\leq} 2\big\| \nabla f_i(\mathbf{x}_t^{(i)}) -\nabla f(\mathbf{x}_t) \big\|^2 + 2\Big\| \frac{1}{N}\sum_{i'=1}^N \big[ \nabla f_{i'}(\mathbf{x}_t^{(i')}) -\sum_{j=1}^N q_t^{j\rightarrow i}\nabla f_j(\mathbf{x}_t^{(j)})\big]\Big\|^2 \\
    &\overset{(b)}{\leq} 2\zeta^2 + \frac{2}{N}\sum_{i'=1}^N \Big\| \nabla f_{i'}(\mathbf{x}_t^{(i')}) -\sum_{j=1}^N q_t^{j\rightarrow i}\nabla f_j(\mathbf{x}_t^{(j)}) \Big\|^2 ,
\end{align*}}
where (a) uses $(\|\mathbf{z}_1+\mathbf{z}_2\|^2)/2 \leq \|\mathbf{z}_1\|^2+\|\mathbf{z}_2\|^2$; (b) is from the definition of $\zeta^2$ in Assumption \ref{assmp:zeta} on the first term and Jensen's inequality on the second term. By rearranging the second term of the right side of the inequality, we get
\allowdisplaybreaks{\begin{align*}
    &\Big(1-\frac{2}{N}\Big) \Big\| \nabla f_i(\mathbf{x}_t^{(i)}) - \sum_{j\in\mathcal{U}} q_t^{j\rightarrow i}\nabla f_j(\mathbf{x}_t^{(j)})\Big\|^2\\
    &\leq 2\zeta^2 +\frac{2}{N} \sum_{i'\in\mathcal{U}\setminus\{i\}} \Big\| \nabla f_{i'}(\mathbf{x}_t^{(i')}) -\sum_{j=1}^N q_t^{j\rightarrow i}\nabla f_j(\mathbf{x}_t^{(j)}) \Big\|^2 \\
    &\leq 2\zeta^2 +\frac{2}{N} \sum_{i'=1}^N \Big\| \nabla f_{i'}(\mathbf{x}_t^{(i')}) -\sum_{j=1}^N q_t^{j\rightarrow i'}\nabla f_j(\mathbf{x}_t^{(j)}) \Big\|^2 .
\end{align*}}
With another rearrangement to the left side, the inequality becomes
\allowdisplaybreaks{\begin{align*}
    \Big(1-\frac{4}{N}\Big) \Big\| \nabla f_i(\mathbf{x}_t^{(i)}) - \sum_{j\in\mathcal{U}} q_t^{j\rightarrow i}\nabla f_j(\mathbf{x}_t^{(j)})\Big\|^2 \leq 2\zeta^2.
\end{align*}}    
\end{proof}

Considering all the above assumptions, we obtain an upper bound on the expectation of the original objective’s gradient when $\mathcal{Q}$ is given in advance.

\begin{theorem}\label{thm:main}
    Let $\mathcal{F}:=f(\mathbf{x}_0)-\min_\mathbf{x}f(\mathbf{x})$. Under all the aforementioned assumptions, we have
    \begin{alignb}
        \min_t \mathbb{E} \bigl[  \| \nabla f(\mathbf{x}_t)\|^2  \big| \mathcal{Q}\bigr] \leq \mathcal{O}\Big(\frac{\mathcal{F}}{B\gamma \Psi} +\frac{\zeta^2}{N-4} +\sigma^2 +N\zeta^2 +BL^2\gamma^2\sigma^2 +\frac{L\gamma \rho^2\sigma^2}{N\Psi}\Big)
    \end{alignb}
    for $\gamma\leq\frac{1}{8BLN\Psi}$, $N>4$, and $\Psi\geq3$.
\end{theorem}

\noindent\textbf{Remark.}\quad We begin, following a similar approach to \cite{wang20-neurips}, by deriving an inequality rooted in the smoothness of $f_i$. This inequality establishes a connection between two local losses from the same user at different timestamps, namely $f_i(\mathbf{x}_{t_0+P}^{(i)})$ and $f_i(\mathbf{x}_{t_0}^{(i)})$. Within this inequality, an inner product term unfolds into several components. Notably, it comprises three distinct subterms: one involving $\|\mathbf{y}_{t,b}^{(j)}-\mathbf{x}_t^{(j)}\|^2$ (refer to Lemma \ref{lemma:ytbj_xtj}), another featuring $\|\mathbf{x}_t^{(j)}-\mathbf{x}_{t_0}^{(j)}\|^2$ (see Lemma \ref{lemma:xt0j_xtj}) which is mainly derived from Algorithm \ref{alg:draco-userview}, and a third term with $\|\nabla f_i(\mathbf{x}_t^{(i)}) - \nabla f_j(\mathbf{x}_t^{(j)}) \|^2$, of which the expectation has an upper bound (refer to Lemma \ref{lemma:dev_local_grads}).
Our proof is novel in the sense that it effectively converts and simplifies the terms on the continuous timeline into discrete values.
Detailed proof is available in Appendix \ref{appendix:proof}. \hfill$\square$

%==================== EXPERIMENTS =========================
\section{Experimental Results}\label{sect:exp}

We conducted experiments with federated learning on two datasets: (1) balanced EMNIST~\cite{cohen17:emnist} dataset with $47$ class labels for image classification tasks, and (2) the Poker hand dataset~\cite{dataset:pokerhand} for multi-class classification tasks, which is widely applied in automatic rule generation.
Each user possesses $1000$ local training samples arranged into training batches with $64$ samples per batch. The default number of participants in each simulation is $N=25$, otherwise it is specified accordingly. The sampling interval is $500$ events, i.e., the evaluation of each local model is done under a test set whenever the 500\textsuperscript{th} event is finished. The rate parameter of exponential distribution in local gradient computation is $\lambda_i=0.1$ for all users by default. In this study, the impact on model compression is not evaluated, implying that the packet size is as large as the raw model. The convolutional neural network (CNN) architecture used in the simulations takes up $596776$ B ($0.57$ MB) for feeding samples from EMNIST, and $51640$ B ($0.05$ MB) from Poker hand, respectively. These values are used to quantify the message size.

\begin{figure*}[t]
    \centering
    \subfloat[\label{subfig:plot-accf1-emnist}]{%
       \includegraphics[width=0.4\textwidth]{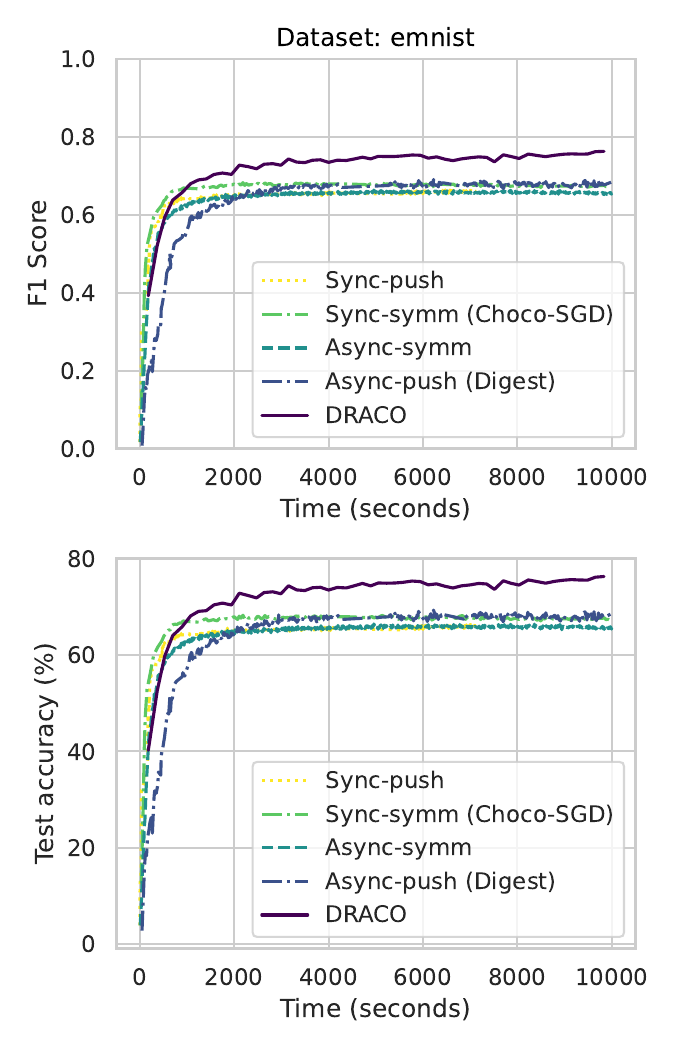}}
    \hspace{1em}
  \subfloat[\label{subfig:plot-accf1-pokerhand}]{%
    \includegraphics[width=0.4\textwidth]{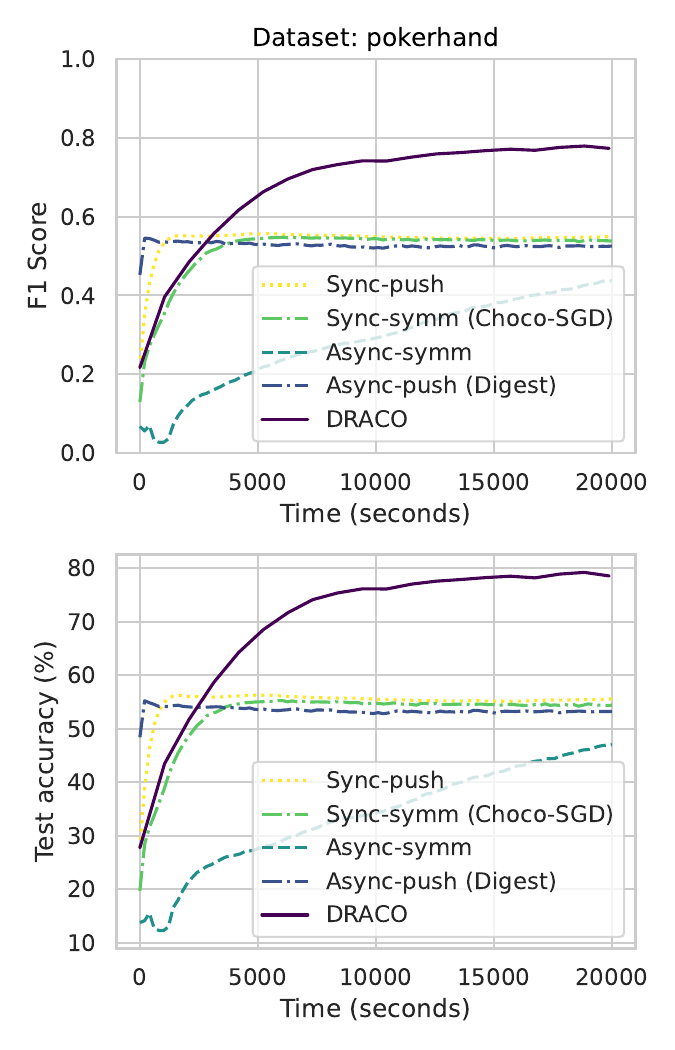}}
  \caption{Performance comparison with the literature under (a) EMNIST dataset, and (b) Poker hand dataset.} %Results for different (a) network topologies ($\Gamma_\text{max}=0.5$); (b) upper bounds on the number of received messages per user. ($P=500$, $\Gamma_\text{max}=10$)
  \label{fig:accf1} 
\end{figure*}

\begin{figure*}
    \centering
    \subfloat[\label{subfig:plot-psi-emnist}]{
        \includegraphics[width=0.4\textwidth]{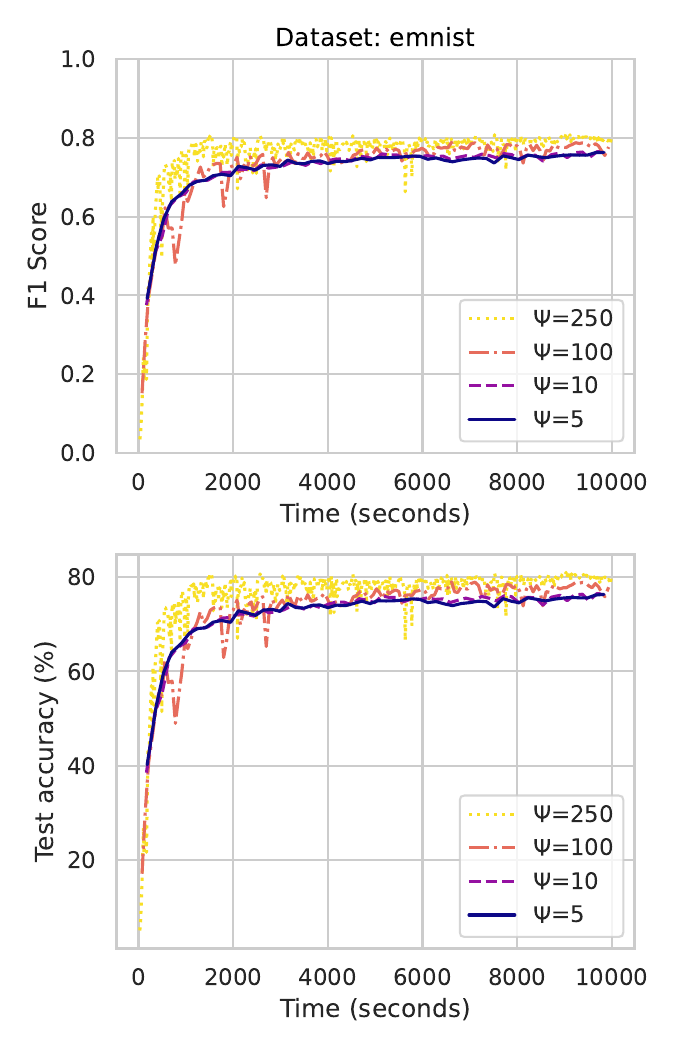}}
    \hspace{1em}
    \subfloat[\label{subfig:plot-psi-pokerhand}]{
        \includegraphics[width=0.4\textwidth]{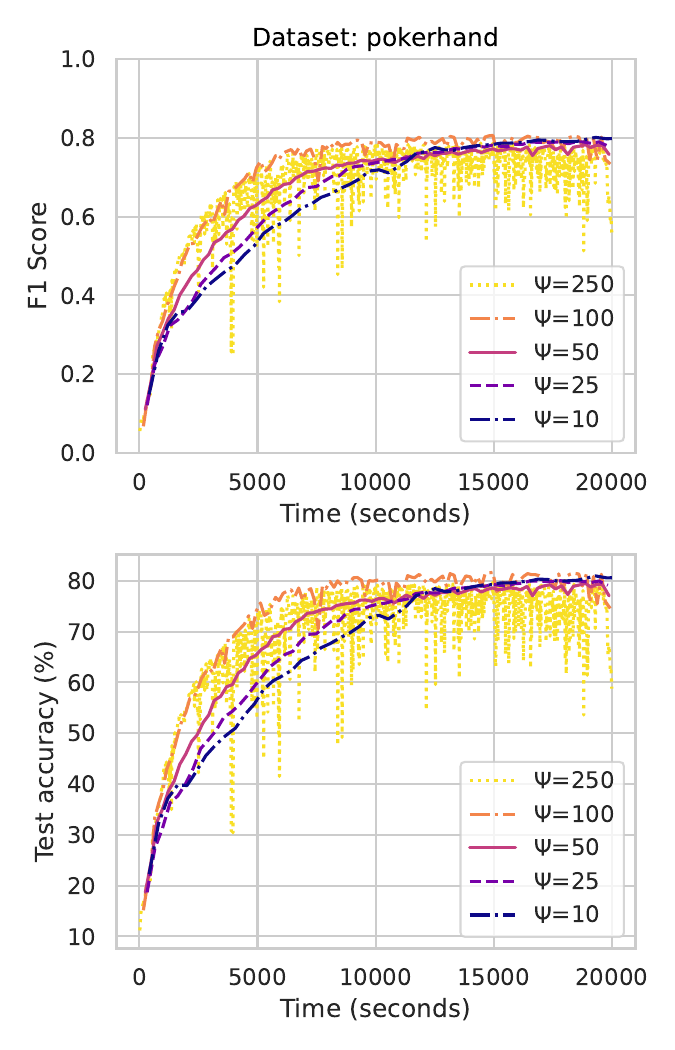}}
    \caption{Results for different upper bounds on the number of received messages per user. ($\Gamma_\text{max}=10$)}
    \label{fig:plot-psi}
\end{figure*}

We performed simulations using two topologies: cycle and complete, with a time-invariant $\mathcal{Q}$. %scenarios according to the time-variability of $\mathcal{Q}$.
The connectivity graph is fixed throughout the whole collaboration process. Each user, indexed $i$ without losing generality, spends some time computing local gradient following $exp(\lambda_i)$ as mentioned in Assumption \ref{assmp:time_comp}. Whenever a local update is done at $t$, user $i$ sends $\Delta_t^{(i)}$ to its neighbors $j\in\mathcal{N}(i)$, where $\mathcal{N}(i)$ indicates a set of user $i$'s neighbors. %Slightly after that moment (i.e., at time $t+t_\epsilon$), $q_{t+t_\epsilon}^{ij}=1$ for all $j$.
Although a pre-defined topology outlines the intended communication paths between nodes, the inherent unreliability of wireless channels can significantly affect data transmission. Factors such as fading, interference, and physical obstructions can disrupt connectivity, resulting in packet losses and delays, thereby undermining the efficiency and reliability of communication within the network.
We used parameters reported in \cite{salehi21} and \cite{xie24} for the wireless communication settings. The radius of the field where the nodes can be scattered is $R=500$ m. We fix the transmit power of each user as $P_i=30$ dBm ($1000$ mW). We also set the path loss exponent $\alpha=4$, the bandwidth $W=10$ MHz, and the noise power density $N_0=-174$ dBm/Hz. We assumed that two nodes interfere with each other during transmission if their distance is closer than $0.1R$.
Due to those wireless communication characteristics, the mechanism for realizing DRACO is slightly different. Specifically, when user $i$ has performed local training at time $t$, it broadcasts its update $\Delta_t^{(i)}$ to all $j\in\mathcal{U}\setminus\{i\}$. It takes
\[ \Gamma_{ij}=\frac{\text{message size}}{W\cdot \log_2(1+\text{SINR}_{i,j})} + \frac{\text{distance}(i,j)}{\text{lightspeed}} \] seconds for the message to arrive at node $j$. Here, the signal-to-interference-plus-noise ratio (SINR) between the two nodes is defined as \[ \text{SINR}_{i,j}=\frac{P_i h_{ji} \text{distance}(j,i)^{-\alpha}}{\sum_{n\in\Phi_j} P_i h_{jn}\text{distance}(j,n)^{-\alpha}+z^2}\ ,\] where $h_{ji}\sim exp(1)$ denotes the small-scale fading gain, $\Phi_j$ is a set of nodes interfering node $j$, and $z^2$ characterizes the variance of AWGN (Additive White Gaussian Noise). As long as the transmission duration $\Gamma_{ij}$ is shorter than the predetermined threshold $\Gamma_\text{max}$, user $j$ succeeds to receive $\Delta^{(i)}$ at time $t+\Gamma_{ij}$. (i.e., $q_{t+\Gamma_{ij}}^{ij}=1$.)

The performance of DRACO is evaluated across different network topologies and datasets. For EMNIST, a cycle topology is employed, where each user is connected to two neighbors. In contrast, the Poker hand dataset utilizes a fully connected topology, with each user directly connected to all others. DRACO's performance is compared against four benchmark methods:
\begin{itemize}
    \item sync-symm: Synchronous learning with symmetric connectivity (Choco-SGD~\cite{koloskova19:chocosgd})
    \item sync-push: Synchronous learning with directed connectivity.
    \item async-symm: Asynchronous learning with symmetric connectivity (Decentralized Asynchronous SGD~\cite{jeong22-async}).
    \item async-push: Asynchronous learning with directed connectivity (Digest~\cite{gholami24:digest}).
\end{itemize}

The term ``Push'' denotes the use of the push-sum algorithm for directed graphs.

The Poker hand dataset presents a unique challenge due to its imbalanced class distribution. To comprehensively assess model performance, both test accuracy and F1-score were evaluated, the latter accounting for both precision and recall.

While the choice of dataset had a minor impact on overall trends, the network topology significantly influenced performance. In the cycle topology, where each user exchanges information with only two neighbors, unreliable channels (e.g., due to fading) can lead to frequent client isolation. As shown in Fig.~\ref{subfig:plot-accf1-emnist}, synchronous methods exhibited comparable performance, but async-symm underperformed async-push, despite using a doubly stochastic matrix. This highlights the sensitivity of async-symm to strict transmission deadlines, emphasizing the importance of well-designed scheduling in asynchronous learning.

In the fully connected topology in Fig.~\ref{subfig:plot-accf1-pokerhand}, where every user is connected to all others, the virtual global model can be trained more robustly, even when some edges are intermittently disrupted. While convergence speeds vary, all algorithms ultimately achieve similar performance.

DRACO consistently outperformed competitors in both test accuracy and F1-score. This advantage stems from its parallel aggregation and unification mechanisms, which effectively mitigate the divergence of local models common in asynchronous decentralized learning. DRACO periodically unifies local reference models and regulates the number of received messages, enhancing robustness in continuous operation and fading environments.

During implementation, performance oscillations were observed when users received excessive redundant updates due to high transmission frequencies (large $\Psi$ values in Fig.~\ref{subfig:plot-psi-emnist} and \ref{subfig:plot-psi-pokerhand}). Conversely, excessively small $\Psi$ values slowed learning by limiting crucial updates' reception. These findings align with prior work \cite{hashemi20} and the theoretical analysis presented in Theorem~\ref{thm:main}.

% =============== CONCLUSION =============================
\section{Concluding Remarks and Future Directions}\label{sect:conc}
We have studied decentralized asynchronous learning optimization through row-stochastic gossip communication networks and proposed a novel method called DRACO. Our technique facilitates the learning process, obviating the need for global iteration counts. It presents local user performance defined on a continuous timeline. We provided practical instructions for each participant by decoupling training and transmission schedules, resulting in complete autonomy and simplified implementations in real-world applications. We analyzed the algorithm convergence and provided experimental results that support the efficacy and feasibility of the proposed framework.

%\subsection{Open Problems}\label{subsect:openprob}
In the remainder, we highlight some promising yet challenging directions that require further investigation.
\begin{itemize}
    \item \textbf{Bandwidth allocation.} In this paper, bandwidth is equally distributed to all users. If the users exchange their SINR information, as well as their weight updates, a bandwidth allocation algorithm can be added within the ``for $i$'' loop, as proposed in \cite{xie24}.
    \item \textbf{More realistic experiments with aggregation time threshold.}
    We can consider that each user has a predetermined threshold to aggregate its neighbors' local updates. The user can perform superposition to its local reference model only after the timeout occurs. For instance, each user $j$ might have an upper bound on the number of $\Delta^{(i)}$'s that it can accept during its receiving period.
    \item \textbf{Improve robustness against collisions.} Random access is known to have a higher probability of collision occurrence. However, it is cumbersome or impractical to predetermine the communication schedule because while carrying out DRACO, the participants decide whether to transmit and/or train their local models without communication or agreement with the other users. Collision in a random access protocol, such as in the context of federated learning where clients transmit messages, can be alleviated by adapting classical approaches in wireless networks. These approaches include configuring a random backoff time after a collision for retransmission attempts, adopting collision detection mechanisms, or allowing clients to dynamically adjust the size of their messages or the transmission power. On the other hand, considering collisions from the resource allocation perspective, the system can assign different priority levels to clients based on factors, such as their data urgency or historical collision rates. 
    \item \textbf{Reception control} We manually selected the rate parameters for transmissions ($\lambda_{ji}$) because we assumed that the participants are not able to predict the frequency of message-receiving events, even in fixed $\mathcal{Q}$ cases. Nevertheless, there exist techniques that enable edge devices to roughly estimate in advance the ratio of successful message reception. With this in mind, it will be possible to study how to manage the reception events in realizing DRACO.
\end{itemize}

%\subsection{Future work}
Future work could also include exploring how the system handles older or outdated updates, which could make it more reliable and efficient. We can also consider using different learning rates or adjustments across various devices, which could make the algorithm work better over a range of device capabilities. Additionally, one can adapt DRACO for mobility scenarios, where distances and communication paths change over time in a three-dimensional space. This could make our approach more practical for real-world applications. Finally, the instructions can be simplified, which would make it easier to use and more accessible in different settings. Addressing the aforementioned challenges could further enhance DRACO's performance and usefulness, opening up new research avenues in asynchronous decentralized federated learning.

\phantomsection
\addcontentsline{toc}{section}{References}
\bibliographystyle{unsrt}  %  plain, unsrt, ieeetr
\small{\bibliography{ref}}

\normalsize

\pagebreak

%=============================================================
\phantomsection
\addcontentsline{toc}{section}{Appendix}
\section*{APPENDICES}
\appendix
\counterwithin*{equation}{section}
\renewcommand{\theequation}{\thesection.\arabic{equation}}
\renewcommand{\thelemma}{\thesection.\arabic{lemma}}
\renewcommand{\theproposition}{\thesection.\arabic{proposition}}
\renewcommand{\thefigure}{\thesection.\arabic{figure}}
\renewcommand{\thedefinition}{\thesection.\arabic{definition}}
\renewcommand{\theassumption}{\thesection.\arabic{assumption}}

\section{Proofs}\label{appendix:proofs}

\begin{figure}[ht]
    \centering
    \includegraphics[width=0.7\textwidth]{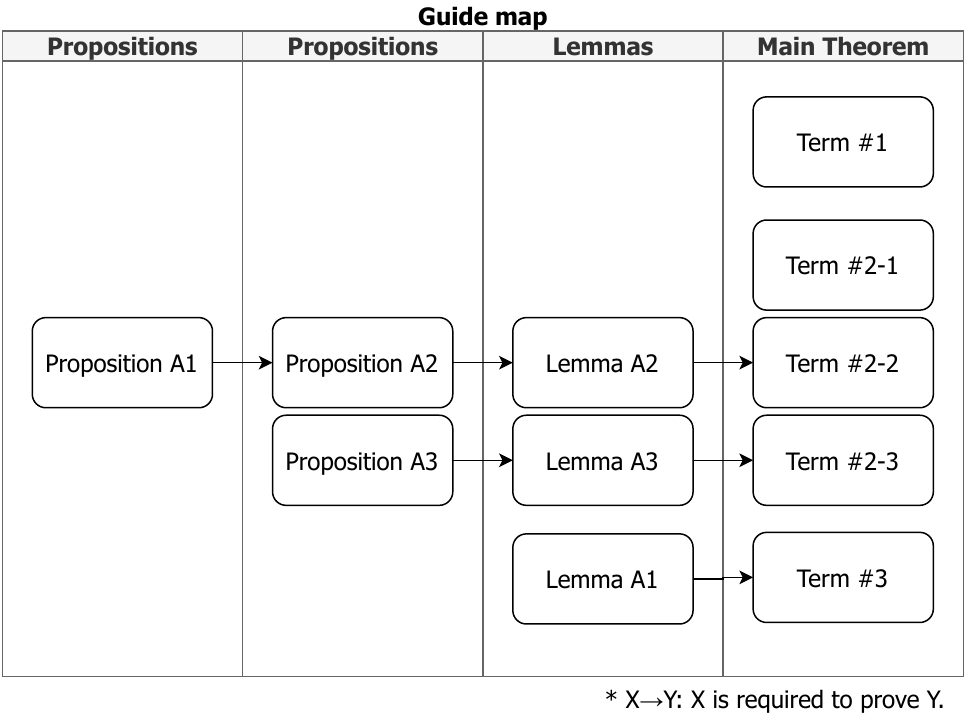}
    \caption{A metaphoric map that guides the correlation of each proposition and lemma for proving the main theorem.}
    \label{doodle:proofmap}
\end{figure}

\subsection{Preliminaries}\label{subsect:prelim}

Before the proof of Theorem \ref{thm:main}, it is essential to verify (i) how many communication events and (ii) how many local gradient updates occur during $P$.
In PPP, communication events occur $\lambda_iP$ times on average during $P$, which indicates the expectation of broadcasting frequency of node $i$. In order to find the bound for $\mathbf{x}_{t_0+P}-\mathbf{x}_{t_0}$, we need to specify how many reception events happen in a random node $i$ during the elapsed time of $P$.
For simplicity, we write $\int_{P}$ to indicate $\int_{t_0}^{t_0+P}$.
The reference model of node $i$ update during $P$ is
\allowdisplaybreaks{\begin{align*}
    \mathbf{x}_{t_0+P}^{(i)}-\mathbf{x}_{t_0}^{(i)} &=
    \int_{P} \sum_j \Pr [i\in \mathcal{N}_t(j)] \Delta_t^{(j)} \dd{t}\\
    &= \int_{P} \sum_j q_t^{j\rightarrow i} \Delta_t^{(j)} \dd{t}\\
    &= \gamma  \int_{P} \sum_j q_t^{j\rightarrow i} \sum_{b=0}^{B-1} \mathbf{g}_j(\mathbf{y}_{\lfloor t \rfloor, b}^{(j)}) \dd{t},
\end{align*}}
which is heterogeneous across nodes. A floored notation $\lfloor t\rfloor$ indicates the latest moment no later than time $t$ that user $j$ computes $\Delta^{(j)}$.

A superscripted or subscripted $\star$ on some variables is analogous to a ``don't-care'' (DC) term in digital logic \cite{karnaugh53:dont_care_term}. For instance, $q_\star$ is the same as any $q_i$, where $i$ can be any user index in $\mathcal{U}$ without loss of generality.

\subsection{Propositions}\label{subsect:prop}
\begin{proposition}{}\label{prop:zeta}
    If Assumption \ref{assmp:zeta} is satisfied, a decentralized learning network with $N\geq 4$ clients satisfies
    \begin{align*}
        \sum_{j\neq i}q_t^{j\rightarrow i} \big\| \nabla f_j(\mathbf{x}_t^{(j)}) - \nabla f_i(\mathbf{x}_t^{(i)}) \big\|^2 \leq \frac{9N\zeta^2}{4}
    \end{align*}
    for all $i,j\in\mathcal{U}$ and $t\in[0,T)$.
\end{proposition}

\begin{proof}
\allowdisplaybreaks{\begin{align*}
    &\sum_{j\neq i}q_t^{j\rightarrow i} \big\| \nabla f_j(\mathbf{x}_t^{(j)}) - \nabla f_i(\mathbf{x}_t^{(i)}) \big\|^2 \\
    &=\sum_{j\neq i}q_t^{j\rightarrow i} \big\| \nabla f_j(\mathbf{x}_t^{(j)}) - \nabla f(\mathbf{x}_t) - \nabla f_i(\mathbf{x}_t^{(i)}) + \nabla f(\mathbf{x}_t) \big\|^2 \\
    &\overset{(a)}{\leq} \Big(1+\frac{1}{\sqrt{N}}\Big) \sum_{j\neq i}q_t^{j\rightarrow i} \big\| \nabla f_j(\mathbf{x}_t^{(j)}) - \nabla f(\mathbf{x}_t) \big\|^2 + (\sqrt{N}+1)\sum_{j\neq i}q_t^{j\rightarrow i}\big\| \nabla f_i(\mathbf{x}_t^{(i)}) - \nabla f(\mathbf{x}_t) \big\|^2 \\
    &\overset{(b)}{\leq} \Big(1+\frac{1}{\sqrt{N}}\Big) \sum_{j\neq i}q_t^{j\rightarrow i} \big\| \nabla f_j(\mathbf{x}_t^{(j)}) - \nabla f(\mathbf{x}_t) \big\|^2 + (\sqrt{N}+1)\zeta^2 \\
    &\overset{(c)}{\leq} (N+2\sqrt{N}+1)\zeta^2
    \quad\overset{(d)}{\leq} \frac{9N\zeta^2}{4},
\end{align*}}
where (a) is due to Young's inequality; (b) comes from Assumption \ref{assmp:zeta}; (c) takes the fact that $q_\star^\star\leq1$ for any user nodes; (d) is always true for $N\geq4$ since $\frac{5N}{4}-2\sqrt{N}-1\geq0$ for any $\sqrt{N}\geq 2$, which satisfies the given condition about $N$.
\end{proof}
\noindent\textbf{Remark.}\quad This proposition appears in the proof of Proposition \ref{prop:ytbj_xtj}.

%\subsubsection{Prop A2}\label{subsubsect:term_2-1}
\begin{proposition}{(Upper bound for superpositioned model deviations.)} \label{prop:ytbj_xtj}
    Let $h_j(b) = \mathbf{y}_{t,b}^{(j)} - \mathbf{x}_t^{(j)}$ denote the difference between a local model calculated by feeding each batch with an index $b$ and the local reference model. For all $i,j\in\mathcal{U}$, when $\gamma\leq\frac{1}{8BL}$, we have
    \begin{align*}
        \sum_{j\neq i} q_t^{j\rightarrow i} \mathbb{E}_{\cdot|\mathcal{Q}}\big[ \| \mathbf{y}_{t,b}^{(j)} - \mathbf{x}_t^{(j)}\|^2 \big] &\leq \frac{2}{5}\sum_{j\neq i}q_t^{j\rightarrow i}\mathbb{E}_{\cdot|\mathcal{Q}} \big[\big\| \mathbf{x}_t^{(i)}-\mathbf{x}_{t_0}^{(i)} \big\|^2\big]\\
            &\quad  +\frac{9N\zeta^2}{10L^2} +\frac{16B\gamma^2\sigma^2}{5} +\frac{128B^2\gamma^2}{5}\mathbb{E}_{\cdot|\mathcal{Q}}[\| \nabla f_i(\mathbf{x}_{t_0}^{(i)})\|^2] \ .
    \end{align*}
\end{proposition}

\begin{proof}
We rephrase the $b+1$\textsuperscript{th} term, $h_j(b+1)=\mathbf{y}_{t,b+1}^{(j)}-\mathbf{x}_t^{(j)}$, as follows.
\allowdisplaybreaks{\begin{alignb}
    &\sum_{j\neq i}q_t^{j\rightarrow i} \mathbb{E}_{\cdot|\mathcal{Q}} \Big[\Big\| \mathbf{y}_{t,b+1}^{(j)}-\mathbf{x}_t^{(j)} \Big\|^2\Big] \\
    &=\sum_{j\neq i}q_t^{j\rightarrow i} \mathbb{E}_{\cdot|\mathcal{Q}} \Big[\Big\|  \mathbf{y}_{t,b}^{(j)}-\mathbf{x}_t^{(j)} -\gamma g_j(\mathbf{y}_{t,b}^{(j)}) \Big\|^2\Big] \\
    % definition of variance
    &\overset{(a)}{=}\sum_{j\neq i}q_t^{j\rightarrow i} \Big\| \mathbb{E}_{\cdot|\mathcal{Q}} \Big[ \mathbf{y}_{t,b}^{(j)} -\mathbf{x}_t^{(j)} -\gamma g_j(\mathbf{y}_{t,b}^{(j)}) \Big]\Big\|^2\\
        &\quad+ \sum_{j\neq i}q_t^{j\rightarrow i}\mathbb{E}_{\cdot|\mathcal{Q}} \Bigg[\Big\| \mathbf{y}_{t,b}^{(j)}-\mathbf{x}_t^{(j)} -\gamma g_j(\mathbf{y}_{t,b}^{(j)}) - \mathbb{E}_{\cdot|\mathcal{Q}}\Big[ \mathbf{y}_{t,b}^{(j)} -\mathbf{x}_t^{(j)} -\gamma g_j(\mathbf{y}_{t,b}^{(j)}) \Big] \Big\|^2\Bigg]\\
    &=\sum_{j\neq i}q_t^{j\rightarrow i} \mathbb{E}_{\cdot|\mathcal{Q}} \Big[\Big\| \mathbf{y}_{t,b}^{(j)} -\mathbf{x}_t^{(j)} -\gamma\nabla f_j(\mathbf{y}_{t,b}^{(j)}) \Big\|^2\Big]   + \sum_{j\neq i}q_t^{j\rightarrow i}\mathbb{E}_{\cdot|\mathcal{Q}} \Big[\Big\| \gamma\big(g_j(\mathbf{y}_{t,b}^{(j)})-\nabla f_j(\mathbf{y}_{t,b}^{(j)})\big) \Big\|^2\Big]\\
    % (b) last term solved   
    &\overset{(b)}{\leq} \sum_{j\neq i}q_t^{j\rightarrow i} \mathbb{E}_{\cdot|\mathcal{Q}} \Big[\Big\| \mathbf{y}_{t,b}^{(j)} -\mathbf{x}_t^{(j)} -\gamma\nabla f_j(\mathbf{y}_{t,b}^{(j)}) \Big\|^2\Big] +\gamma^2\sigma^2 \\
    % adding-subtracting \nabla f_j(x_t^j), \nabla f_i(x_t^i), and \nabla f_i(x_t0^i)
    &=\sum_{j\neq i}q_t^{j\rightarrow i} \mathbb{E}_{\cdot|\mathcal{Q}} \Big[\Big\| \mathbf{y}_{t,b}^{(j)} -\mathbf{x}_t^{(j)} + \gamma\nabla f_j(\mathbf{x}_t^{(j)}) -\gamma\nabla f_j(\mathbf{y}_{t,b}^{(j)}) - \gamma\nabla f_j(\mathbf{x}_t^{(j)}) -\gamma\nabla f_i(\mathbf{x}_t^{(i)})\\
        &\hspace{7em}+\gamma\nabla f_i(\mathbf{x}_t^{(i)}) -\gamma\nabla f_i(\mathbf{x}_{t_0}^{(i)}) +\gamma\nabla f_i(\mathbf{x}_{t_0}^{(i)}) \Big\|^2\Big] +\gamma^2\sigma^2\\
    % (c) Young's inequality
    &\overset{(c)}{\leq} \Big(1+\frac{1}{2B-1}\Big) \sum_{j\neq i}q_t^{j\rightarrow i} \mathbb{E}_{\cdot|\mathcal{Q}} \big[\big\| \mathbf{y}_{t,b}^{(j)}-\mathbf{x}_t^{(j)} \big\|^2\big]\\
        &\quad+2B\gamma^2 \sum_{j\neq i}q_t^{j\rightarrow i} \mathbb{E}_{\cdot|\mathcal{Q}} \Big[\Big\| \nabla f_j(\mathbf{y}_{t,b}^{(j)})-\nabla f_j(\mathbf{x}_t^{(j)}) + \nabla f_j(\mathbf{x}_t^{(j)}) -\nabla f_i(\mathbf{x}_t^{(i)}) \\
        &\hspace{11em}+\nabla f_i(\mathbf{x}_t^{(i)}) -\nabla f_i(\mathbf{x}_{t_0}^{(i)}) +\nabla f_i(\mathbf{x}_{t_0}^{(i)})\Big\|^2\Big]  +\gamma^2\sigma^2\\
    % split into four terms 
    &\leq \Big(1+\frac{1}{2B-1}\Big) \sum_{j\neq i}q_t^{j\rightarrow i} \mathbb{E}_{\cdot|\mathcal{Q}} \big[\big\| \mathbf{y}_{t,b}^{(j)}-\mathbf{x}_t^{(j)} \big\|^2\big]\\
        &\quad+8B\gamma^2 \sum_{j\neq i}q_t^{j\rightarrow i} \mathbb{E}_{\cdot|\mathcal{Q}} \Big[\Big\| \nabla f_j(\mathbf{y}_{t,b}^{(j)})-\nabla f_j(\mathbf{x}_t^{(j)}) \Big\|^2\Big] \\
        &\quad+8B\gamma^2 \sum_{j\neq i}q_t^{j\rightarrow i} \mathbb{E}_{\cdot|\mathcal{Q}} \big[\big\| \nabla f_j(\mathbf{x}_t^{(j)}) - \nabla f_i(\mathbf{x}_t^{(i)}) \big\|^2\big] \\
        &\quad+8B\gamma^2 \sum_{j\neq i}q_t^{j\rightarrow i} \mathbb{E}_{\cdot|\mathcal{Q}} \big[\big\| \nabla f_i(\mathbf{x}_t^{(i)}) -\nabla f_i(\mathbf{x}_{t_0}^{(i)}) \big\|^2\big] \\
        &\quad+8B\gamma^2 \sum_{j\neq i}q_t^{j\rightarrow i} \mathbb{E}_{\cdot|\mathcal{Q}} \big[\big\| \nabla f_i(\mathbf{x}_{t_0}^{(i)}) \big\|^2\big] +\gamma^2\sigma^2\\
    % (d) L-smoothness for 2rd and 4th term
    % proposition (sub-lemma) 1 for 3rd term
    &\overset{(d)}{\leq} \Big(1+\frac{1}{2B-1}\Big) \sum_{j\neq i}q_t^{j\rightarrow i} \mathbb{E}_{\cdot|\mathcal{Q}} \big[\big\| \mathbf{y}_{t,b}^{(j)} -\mathbf{x}_t^{(j)} \big\|^2\big]   +8BL^2\gamma^2 \sum_{j\neq i}q_t^{j\rightarrow i} \mathbb{E}_{\cdot|\mathcal{Q}} \big[\big\| \mathbf{y}_{t,b}^{(j)} -\mathbf{x}_t^{(j)} \big\|^2\big] \\ 
        &\quad+18BN\gamma^2\zeta^2 + 8BL^2\gamma^2\sum_{j\neq i}q_t^{j\rightarrow i} \mathbb{E}_{\cdot|\mathcal{Q}} \big[\big\| \mathbf{x}_t^{(i)}-\mathbf{x}_{t_0}^{(i)} \big\|^2\big] +8B\gamma^2  \mathbb{E}_{\cdot|\mathcal{Q}} [\| \nabla f_i(\mathbf{x}_{t_0}^{(i)}) \|^2]  +\gamma^2\sigma^2\\
    % integrated terms ending with |h(b)|^2
    % integrated terms ending with \gamma^2\sigma^2
    &= \Big(1+8BL^2\gamma^2+\frac{1}{2B-1}\Big) \sum_{j\neq i}q_t^{j\rightarrow i} \mathbb{E}_{\cdot|\mathcal{Q}} \big[\| \mathbf{y}_{t,b}^{(j)} -\mathbf{x}_t^{(j)} \|^2\big] \\
        &\quad+18BN\gamma^2\zeta^2 +\gamma^2\sigma^2 + 8BL^2\gamma^2\sum_{j\neq i}q_t^{j\rightarrow i} \mathbb{E}_{\cdot|\mathcal{Q}} \big[\big\| \mathbf{x}_t^{(i)}-\mathbf{x}_{t_0}^{(i)} \big\|^2\big]+8B\gamma^2\mathbb{E}_{\cdot|\mathcal{Q}}[\| \nabla f_i(\mathbf{x}_{t_0}^{(i)})\|^2] \\
    &\overset{(e)}{\leq} \Bigg(1+\frac{5}{8\big(B-\frac{1}{2}\big)}\Bigg) \sum_{j\neq i}q_t^{j\rightarrow i} \mathbb{E}_{\cdot|\mathcal{Q}} \big[\| \mathbf{y}_{t,b}^{(j)} -\mathbf{x}_t^{(j)} \|^2\big] +\frac{9N\zeta^2}{32BL^2} +\gamma^2\sigma^2 \\
        &\quad+\frac{1}{8B}\sum_{j\neq i}q_t^{j\rightarrow i}\mathbb{E}_{\cdot|\mathcal{Q}} \big[\big\| \mathbf{x}_t^{(i)}-\mathbf{x}_{t_0}^{(i)} \big\|^2\big] +8B\gamma^2\mathbb{E}_{\cdot|\mathcal{Q}}[\|\nabla f_i(\mathbf{x}_{t_0}^{(i)})\|^2]
    \label{eq:term2.1}
\end{alignb}}
where (a) is from the definition of variance; (b) is derived from the definition of $\sigma$ in Assumption \ref{assmp:sg}; (c) Young's inequality; (d) uses $L$-smoothness on the second and the fourth term, and applies Proposition \ref{prop:zeta} on the third term. Afterwards, the first two terms are integrated; (e) is derived from the fact that $\gamma^2\leq\frac{1}{64B^2L^2}$ and that
\begin{align*}
    8BL^2\gamma^2 + \frac{1}{2B-1} \leq \frac{1}{8B} + \frac{1}{2B-1} \leq  \frac{1}{8B-4} + \frac{1}{2B-1} = \frac{5}{8\big(B-\frac{1}{2}\big)} .
\end{align*}
%\tgreen{[$\mathbb{E}_{\cdot|\mathcal{Q}} [\| \nabla f(\mathbf{x}_t)\|^2]$ appears again in inequality \ref{eq:thm_3}.]}

Let $H(b)$ indicate $\sum_{j\neq i}q_t^{j\rightarrow i}\mathbb{E}_{\cdot|\mathcal{Q}} \big[\big\| \mathbf{y}_{t,b}^{(j)} -\mathbf{x}_t^{(j)} \big\|^2\big]$. From the last line of inequality \ref{eq:term2.1}, we have
\begin{alignb}
   H(b+1) &\leq \Bigg(1+\frac{5}{8\big(B-\frac{1}{2}\big)}\Bigg) H(b) +\frac{1}{8B}\sum_{j\neq i}q_t^{j\rightarrow i}\mathbb{E}_{\cdot|\mathcal{Q}} \big[\big\| \mathbf{x}_t^{(i)}-\mathbf{x}_{t_0}^{(i)} \big\|^2\big] \\
   &\quad+18BN\gamma^2\zeta^2 +\gamma^2\sigma^2+8B\gamma^2\mathbb{E}_{\cdot|\mathcal{Q}}[\|\nabla f_i(\mathbf{x}_{t_0}^{(i)})\|^2]\ .
   \label{H(b+1)}
\end{alignb}

Since $\mathbf{y}_{t,0}^{(j)} = \mathbf{x}_t^{(j)}$ for all $t,\ j$ based on Algorithm \ref{alg:draco-userview},
\begin{align*}
    H(0) = \sum_{j\neq i}q_t^{j\rightarrow i} \mathbb{E}_{\cdot|\mathcal{Q}}\big[\| \mathbf{y}_{t,0}^{(j)} -\mathbf{x}_t^{(j)} \|^2\big] = 0 .
\end{align*}
Recurring inequality \ref{H(b+1)} from $H(0)$, we can get
\allowdisplaybreaks{\begin{align*}
    H(b) &\leq \Bigg(1+\frac{5}{8\big(B-\frac{1}{2}\big)}\Bigg)^b H(0) + \sum_{b'=0}^{b-1}  \Bigg(1+\frac{5}{8\big(B-\frac{1}{2}\big)}\Bigg)^{b'}\\
    &\quad\cdot \Big( \frac{1}{8B}\sum_{j\neq i}q_t^{j\rightarrow i}\mathbb{E}_{\cdot|\mathcal{Q}} \big[\big\| \mathbf{x}_t^{(i)}-\mathbf{x}_{t_0}^{(i)} \big\|^2\big]+ \frac{9N\zeta^2}{32BL^2} +\gamma^2\sigma^2 +8B\gamma^2\mathbb{E}_{\cdot|\mathcal{Q}}[\| \nabla f_i(\mathbf{x}_{t_0}^{(i)})\|^2] \Big)\\
    % removed H(0) term
    &\leq \Big(\frac{1}{8B}\sum_{j\neq i}q_t^{j\rightarrow i}\mathbb{E}_{\cdot|\mathcal{Q}} \big[\big\| \mathbf{x}_t^{(i)}-\mathbf{x}_{t_0}^{(i)} \big\|^2\big] + \frac{9N\zeta^2}{32BL^2} +\gamma^2\sigma^2 +8B\gamma^2\mathbb{E}_{\cdot|\mathcal{Q}}[\| \nabla f_i(\mathbf{x}_{t_0}^{(i)})\|^2] \Big)\\
        &\quad\cdot \sum_{b=0}^{B-1} \Bigg(1+\frac{5}{8\big(B-\frac{1}{2}\big)}\Bigg)^{b}\\
    &= \Bigg[\Bigg(1+\frac{5}{8\big(B-\frac{1}{2}\big)}\Bigg)^B -1\Bigg] \cdot \frac{8\big(B-\frac{1}{2}\big)}{5} \\
        &\quad\cdot \Big( \frac{1}{8B}\sum_{j\neq i}q_t^{j\rightarrow i}\mathbb{E}_{\cdot|\mathcal{Q}} \big[\big\| \mathbf{x}_t^{(i)}-\mathbf{x}_{t_0}^{(i)} \big\|^2\big] +\frac{9N\zeta^2}{32BL^2} +\gamma^2\sigma^2 +8B\gamma^2\mathbb{E}_{\cdot|\mathcal{Q}}[\| \nabla f_i(\mathbf{x}_{t_0}^{(i)})\|^2] \Big) \\
    &= \Bigg[\Bigg(1+\frac{5}{8\big(B-\frac{1}{2}\big)}\Bigg)^{B-\frac{1}{2}} \Bigg(1+\frac{5}{8\big(B-\frac{1}{2}\big)}\Bigg)^{\frac{1}{2}} -1\Bigg] \cdot \frac{8\big(B-\frac{1}{2}\big)}{5} \\
        &\quad\cdot \Big(\frac{1}{8B}\sum_{j\neq i}q_t^{j\rightarrow i}\mathbb{E}_{\cdot|\mathcal{Q}} \big[\big\| \mathbf{x}_t^{(i)}-\mathbf{x}_{t_0}^{(i)} \big\|^2\big] +\frac{9N\zeta^2}{32BL^2} +\gamma^2\sigma^2  +8B\gamma^2\mathbb{E}_{\cdot|\mathcal{Q}}[\| \nabla f_i(\mathbf{x}_{t_0}^{(i)})\|^2] \Big) \\
    &\overset{(a)}{\leq} \Bigg[e^{\frac{5}{8}} \cdot \frac{3}{2}-1\Bigg] \cdot \frac{8\big(B-\frac{1}{2}\big)}{5}\\
        &\quad\cdot \Big(\frac{1}{8B}\sum_{j\neq i}q_t^{j\rightarrow i}\mathbb{E}_{\cdot|\mathcal{Q}} \big[\big\| \mathbf{x}_t^{(i)}-\mathbf{x}_{t_0}^{(i)} \big\|^2\big] +\frac{9N\zeta^2}{32BL^2} +\gamma^2\sigma^2 +8B\gamma^2\mathbb{E}_{\cdot|\mathcal{Q}}[\| \nabla f_i(\mathbf{x}_{t_0}^{(i)})\|^2] \Big) \\
    &\overset{(b)}{\leq} \frac{16}{5}B \Big(\frac{1}{8B}\sum_{j\neq i}q_t^{j\rightarrow i}\mathbb{E}_{\cdot|\mathcal{Q}} \big[\big\| \mathbf{x}_t^{(i)}-\mathbf{x}_{t_0}^{(i)} \big\|^2\big]  +\frac{9N\zeta^2}{32BL^2} +\gamma^2\sigma^2 +8B\gamma^2\mathbb{E}_{\cdot|\mathcal{Q}}[\| \nabla f_i(\mathbf{x}_{t_0}^{(i)})\|^2] \Big) \\
    % final
    &= \frac{2}{5}\sum_{j\neq i}q_t^{j\rightarrow i}\mathbb{E}_{\cdot|\mathcal{Q}} \big[\big\| \mathbf{x}_t^{(i)}-\mathbf{x}_{t_0}^{(i)} \big\|^2\big]  +\frac{9N\zeta^2}{10L^2} +\frac{16B\gamma^2\sigma^2}{5} +\frac{128B^2\gamma^2}{5}\mathbb{E}_{\cdot|\mathcal{Q}}[\| \nabla f_i(\mathbf{x}_{t_0}^{(i)})\|^2]
\end{align*}}
where (a) comes from $(1+x)^{1/x}\leq e$ and $B\geq 1$, which results in
$\Big(1+\frac{5}{8(B-(1/2))}\Big)^{1/2} \leq \big(1+\frac{5}{4}\big)^{1/2} = \frac{3}{2}$; (b) is due to $\frac{3}{2}e^{\frac{5}{8}} -1 \leq 2$ and $B-\frac{1}{2}\leq B$.
\end{proof}
\noindent\textbf{Remark.}\quad This proposition appears in the proof of Lemma \ref{lemma:ytbj_xtj}, which is used at the first term of inequality \ref{eq:thm_2}.

%\subsubsection{Prop A3}\label{subsubsect:term_2-2}

\begin{proposition}{(Upper bound for the local reference model change)}\label{prop:xt0j_xtj}
    %Let $n_t$ indicate the number of transception events during the given period $[t_0, t)$.
    When Assumption \ref{assmp:finite_Psi} holds and $\gamma\leq\min(\frac{1}{8BL}, \frac{1}{8BLN\Psi})$, we have
    \begin{align*}
        \mathbb{E}_{\cdot|\mathcal{Q}} \Big[ \sum_{j\neq i} q_t^{j\rightarrow i} \big\| \mathbf{x}_t^{(j)} -\mathbf{x}_{t_0}^{(j)} \big\|^2\Big]
        &\leq 2 \mathbb{E}_{\cdot|\mathcal{Q}} \Big[ \sum_{j\neq i} q_t^{j\rightarrow i} \| \mathbf{y}_{t,b}^{(j)}-\mathbf{x}_t^{(j)}\|^2 \Big]+ \frac{8\zeta^2}{L^2(N-4)} \\
            &\quad + \frac{1}{16L^2N^2} \mathbb{E}_{\cdot|\mathcal{Q}} \big[\big\| \nabla f_i(\mathbf{x}_{t_0}^{(i)}) \big\|^2\big]     +\frac{3\sigma^2}{16L^2}\ ,        
        %&\leq 2\mathbb{E}_{\cdot|\mathcal{Q}} \Big[ \sum_{j\neq i} q_t^{j\rightarrow i} \big\|\mathbf{y}_{t,b}^{(j)} - \mathbf{x}_t^{(j)} \big\|^2 \Big] + \frac{4\zeta^2}{L^2(N-4)}\\
        %&\quad +  \frac{1}{L^2N} \mathbb{E}_{\cdot|\mathcal{Q}} \big[\big\| \nabla f_i(\mathbf{x}_{t_0}^{(i)}) \big\|^2\big] +\frac{\sigma^2}{500L^2}\ ,
    \end{align*} for all $i,j\in\mathcal{U}$ and for $t\in[t_0, t_0+P)$.
\end{proposition}

\begin{proof}
Here, we use $\int_{\tau}$ to replace $\int_{\tau=t_0}^t$ for simplicity of writing. The left side of the inequality can be rephrased as below by bringing Appendix \ref{subsect:prelim}.
\allowdisplaybreaks{\begin{alignb}
    &\mathbb{E}_{\cdot|\mathcal{Q}} \Big[ \sum_{j\neq i} q_t^{j\rightarrow i} \big\| \mathbf{x}_t^{(j)} -\mathbf{x}_{t_0}^{(j)} \big\|^2\Big] \\
    &=\mathbb{E}_{\cdot|\mathcal{Q}} \Bigg[ \sum_{j\neq i} q_t^{j\rightarrow i} \Bigg\| \gamma\int_{\tau} \sum_{n\neq j} q_\tau^{n\rightarrow j} \sum_{b=0}^{B-1} g_n(\mathbf{y}_{\tau,b}^{(n)})\dd{\tau} \Bigg\|^2\Bigg]\\
    &=\mathbb{E}_{\cdot|\mathcal{Q}} \Bigg[ \sum_{j\neq i} q_t^{j\rightarrow i} \Bigg\| \gamma\int_{\tau} \sum_{n\neq j} q_\tau^{n\rightarrow j} \sum_{b=0}^{B-1} \nabla f_n(\mathbf{y}_{\tau,b}^{(n)}) \dd{\tau} \\
        &\hspace{8em}+  \gamma\int_{\tau} \sum_{n\neq j} q_\tau^{n\rightarrow j} \sum_{b=0}^{B-1} \big[ g_n(\mathbf{y}_{\tau,b}^{(n)}) -\nabla f_n(\mathbf{y}_{\tau,b}^{(n)}) \big]\dd{\tau} \Bigg\|^2\Bigg]\\
    % first introduction of "\mu_1"
    &\overset{(i)}{\leq} \mu_1 \mathbb{E}_{\cdot|\mathcal{Q}} \Bigg[ \sum_{j\neq i} q_t^{j\rightarrow i} \Bigg\| \gamma\int_{\tau} \sum_{n\neq j} q_\tau^{n\rightarrow j} \sum_{b=0}^{B-1} \nabla f_n(\mathbf{y}_{\tau,b}^{(n)}) \dd{\tau} \Bigg\|^2\Bigg]\\
        &\quad +\Big(1+\frac{1}{\mu_1-1}\Big) \mathbb{E}_{\cdot|\mathcal{Q}} \Bigg[ \sum_{j\neq i} q_t^{j\rightarrow i} \Bigg\| \gamma\int_{\tau} \sum_{n\neq j} q_\tau^{n\rightarrow j} \sum_{b=0}^{B-1} \big[ g_n(\mathbf{y}_{\tau,b}^{(n)}) -\nabla f_n(\mathbf{y}_{\tau,b}^{(n)}) \big]\dd{\tau} \Bigg\|^2\Bigg] \\
    % (a1) sigma in assumption 3
    %      pulling out the variables inside the L2 norms
    &\overset{(a)}{\leq}\mu_1 \mathbb{E}_{\cdot|\mathcal{Q}} \Bigg[ \sum_{j\neq i} q_t^{j\rightarrow i} \Bigg\| \gamma\int_{\tau} \sum_{n\neq j} q_\tau^{n\rightarrow j} \sum_{b=0}^{B-1} \nabla f_n(\mathbf{y}_{\tau,b}^{(n)}) \dd{\tau} \Bigg\|^2\Bigg]\\
        &\quad +\Big(1+\frac{1}{\mu_1-1}\Big)\gamma^2 \mathbb{E}_{\cdot|\mathcal{Q}} \Bigg[ \sum_{j\neq i} q_t^{j\rightarrow i} \psi_j(t_0,t) \int_{/
        \tau} \Big\|\sum_{n\neq j} q_\tau^{n\rightarrow j} \sum_{b=0}^{B-1} \big[ g_n(\mathbf{y}_{\tau,b}^{(n)}) -\nabla f_n(\mathbf{y}_{\tau,b}^{(n)}) \big] \Big\|^2\dd{\tau}\Bigg] \\
    %(a2) q^2 \leq q
    %     pulling out the variables inside the L2 norms
    &\leq \mu_1 \mathbb{E}_{\cdot|\mathcal{Q}} \Bigg[ \sum_{j\neq i} q_t^{j\rightarrow i} \Bigg\| \gamma\int_{\tau} \sum_{n\neq j} q_\tau^{n\rightarrow j} \sum_{b=0}^{B-1} \nabla f_n(\mathbf{y}_{\tau,b}^{(n)}) \dd{\tau} \Bigg\|^2\Bigg]\\
        &\quad +\Big(1+\frac{1}{\mu_1-1}\Big) N\gamma^2 \mathbb{E}_{\cdot|\mathcal{Q}} \Bigg[ \sum_{j\neq i} q_t^{j\rightarrow i} \psi_j(t_0,t) \int_{\tau} \sum_{n\neq j} q_\tau^{n\rightarrow j} \Big\|\sum_{b=0}^{B-1} \big[ g_n(\mathbf{y}_{\tau,b}^{(n)}) -\nabla f_n(\mathbf{y}_{\tau,b}^{(n)}) \big] \Big\|^2\dd{\tau}\Bigg] \\
    %(a3) B out
    %     pulling out the variables inside the L2 norms
    &\leq \mu_1 \mathbb{E}_{\cdot|\mathcal{Q}} \Bigg[ \sum_{j\neq i} q_t^{j\rightarrow i} \Bigg\| \gamma\int_{\tau} \sum_{n\neq j} q_\tau^{n\rightarrow j} \sum_{b=0}^{B-1} \nabla f_n(\mathbf{y}_{\tau,b}^{(n)}) \dd{\tau} \Bigg\|^2\Bigg]\\
        &\quad +\Big(1+\frac{1}{\mu_1-1}\Big) BN\gamma^2 \mathbb{E}_{\cdot|\mathcal{Q}} \Bigg[ \sum_{j\neq i} q_t^{j\rightarrow i} \psi_j(t_0,t) \int_{\tau} \sum_{n\neq j} q_\tau^{n\rightarrow j} \sum_{b=0}^{B-1} \big\| g_n(\mathbf{y}_{\tau,\star}^{(n)}) -\nabla f_n(\mathbf{y}_{\tau,\star}^{(n)}) \big\|^2\dd{\tau}\Bigg] \\
    %&\leq \mu_1 \mathbb{E}_{\cdot|\mathcal{Q}} \Bigg[ \sum_{j\neq i} q_t^{j\rightarrow i} \Bigg\| \gamma\int_{\tau} \sum_{j\neq i} q_\tau^{j\rightarrow i} \sum_{b=0}^{B-1} \nabla f_j(\mathbf{y}_{\tau,b}^{(j)}) \dd{\tau} \Bigg\|^2\Bigg]\\    &\quad +\Big(1+\frac{1}{\mu_1-1}\Big) B^2N\gamma^2\psi_j^2(t_0,t) \mathbb{E}_{\cdot|\mathcal{Q}} \Big[ \sum_{j\neq i} q_t^{j\rightarrow i} \big\| g_j(\mathbf{y}_{\star,\star}^{(j)}) -\nabla f_j(\mathbf{y}_{\star,\star}^{(j)}) \big\|^2 \Big] \\
    &\overset{(b)}{\leq} \mu_1 \mathbb{E}_{\cdot|\mathcal{Q}} \Bigg[ \sum_{j\neq i} q_t^{j\rightarrow i} \Bigg\| \gamma\int_{\tau} \sum_{n\neq j} q_\tau^{n\rightarrow j} \sum_{b=0}^{B-1} \nabla f_n(\mathbf{y}_{\tau,b}^{(n)}) \dd{\tau} \Bigg\|^2\Bigg] +\Big(1+\frac{1}{\mu_1-1}\Big) B^2N^2\gamma^2\psi_j^2(t_0,t)\sigma^2 \\
    % (**) adding-subtracting that resembles (*) in Theorem 1's proof.
    %&\overset{\tred{(**)}}{\leq}
    &\leq \mu_1 \mathbb{E}_{\cdot|\mathcal{Q}} \Bigg[ \sum_{j\neq i} q_t^{j\rightarrow i} \Bigg\|\gamma\int_{\tau} \sum_{n\neq j} q_\tau^{n\rightarrow j} \sum_{b=0}^{B-1} \Big[\nabla f_n(\mathbf{y}_{\tau,b}^{(n)}) - \nabla f_n(\mathbf{x}_\tau^{(n)}) + \nabla f_n(\mathbf{x}_\tau^{(n)}) - \nabla f_j(\mathbf{x}_\tau^{(j)})\\
        &\hspace{4em}+ \nabla f_j(\mathbf{x}_\tau^{(j)}) - \nabla f_j(\mathbf{x}_{t_0}^{(j)}) + \nabla f_j(\mathbf{x}_{t_0}^{(j)}) -\nabla f_i(\mathbf{x}_{t_0}^{(i)}) +\nabla f_i(\mathbf{x}_{t_0}^{(i)}) \Big] \dd{\tau} \Bigg\|^2\Bigg]\\
        &\quad +\Big(1+\frac{1}{\mu_1-1}\Big) B^2N^2\gamma^2\psi_j^2(t_0,t)\sigma^2 \\
    % separation
    &= \mu_1 \mathbb{E}_{\cdot|\mathcal{Q}} \Bigg[ \sum_{j\neq i} q_t^{j\rightarrow i} \Bigg\|\gamma\int_{\tau} \sum_{n\neq j} q_\tau^{n\rightarrow j} \sum_{b=0}^{B-1} \Big[\nabla f_n(\mathbf{y}_{\tau,b}^{(n)}) - \nabla f_n(\mathbf{x}_\tau^{(n)}) \Big] \dd{\tau} \\
        &\hspace{8em} + \gamma\int_{\tau} \sum_{n\neq j} q_\tau^{n\rightarrow j} \sum_{b=0}^{B-1} \Big[\nabla f_n(\mathbf{x}_\tau^{(n)}) - \nabla f_j(\mathbf{x}_\tau^{(j)}) \Big] \dd{\tau}\\
        &\hspace{8em}+ \gamma\int_{\tau} \sum_{n\neq j} q_\tau^{n\rightarrow j} \sum_{b=0}^{B-1} \Big[\nabla f_j(\mathbf{x}_\tau^{(j)}) - \nabla f_j(\mathbf{x}_{t_0}^{(j)}) \Big] \dd{\tau}\\
        &\hspace{8em}+ \gamma\int_{\tau} \sum_{n\neq j} q_\tau^{n\rightarrow j} \sum_{b=0}^{B-1} \Big[\nabla f_j(\mathbf{x}_{t_0}^{(j)}) -\nabla f_i(\mathbf{x}_{t_0}^{(i)}) \Big] \dd{\tau}\\
        &\hspace{8em} + \gamma\int_{\tau} \sum_{n\neq j} q_\tau^{n\rightarrow j} \sum_{b=0}^{B-1} \nabla f_i(\mathbf{x}_{t_0}^{(i)}) \dd{\tau} \Bigg\|^2\Bigg] +\Big(1+\frac{1}{\mu_1-1}\Big) B^2N^2\gamma^2\psi_j^2(t_0,t)\sigma^2 \\
    % (c) Jensen's inequality
    % first introduction of "\mu_2"
    &\overset{(ii, c)}{\leq}  4\mu_1\mu_2 \mathbb{E}_{\cdot|\mathcal{Q}} \Bigg[ \sum_{j\neq i} q_t^{j\rightarrow i} \Bigg\|\gamma\int_{\tau} \sum_{n\neq j} q_\tau^{n\rightarrow j} \sum_{b=0}^{B-1} \Big[\nabla f_n(\mathbf{y}_{\tau,b}^{(n)}) - \nabla f_n(\mathbf{x}_\tau^{(n)}) \Big] \dd{\tau} \Bigg\|^2\Bigg] \\
        &\quad + 4\mu_1\mu_2 \mathbb{E}_{\cdot|\mathcal{Q}} \Bigg[ \sum_{j\neq i} q_t^{j\rightarrow i} \Bigg\| \gamma\int_{\tau} \sum_{n\neq j} q_\tau^{n\rightarrow j} \sum_{b=0}^{B-1} \Big[\nabla f_n(\mathbf{x}_\tau^{(n)}) - \nabla f_j(\mathbf{x}_\tau^{(j)}) \Big] \dd{\tau} \Bigg\|^2\Bigg]\\
        &\quad + 4\mu_1\mu_2 \mathbb{E}_{\cdot|\mathcal{Q}} \Bigg[ \sum_{j\neq i} q_t^{j\rightarrow i} \Bigg\| \gamma\int_{\tau} \sum_{n\neq j} q_\tau^{n\rightarrow j} \sum_{b=0}^{B-1} \Big[\nabla f_j(\mathbf{x}_\tau^{(j)}) - \nabla f_j(\mathbf{x}_{t_0}^{(j)}) \Big] \dd{\tau} \Bigg\|^2\Bigg]\\
        &\quad + 4\mu_1\mu_2 \mathbb{E}_{\cdot|\mathcal{Q}} \Bigg[ \sum_{j\neq i} q_t^{j\rightarrow i} \Bigg\| \gamma\int_{\tau} \sum_{n\neq j} q_\tau^{n\rightarrow j} \sum_{b=0}^{B-1} \Big[\nabla f_j(\mathbf{x}_{t_0}^{(j)}) - \nabla f_i(\mathbf{x}_{t_0}^{(i)}) \Big] \dd{\tau} \Bigg\|^2\Bigg]\\
        &\quad + \mu_1\Big(1+\frac{1}{\mu_2-1}\Big) \mathbb{E}_{\cdot|\mathcal{Q}} \Bigg[ \sum_{j\neq i} q_t^{j\rightarrow i} \Bigg\| \gamma\int_{\tau} \sum_{n\neq j} q_\tau^{n\rightarrow j} \sum_{b=0}^{B-1} \nabla f_i(\mathbf{x}_{t_0}^{(i)}) \dd{\tau} \Bigg\|^2\Bigg]\\
        &\quad+\Big(1+\frac{1}{\mu_1-1}\Big) B^2N^2\gamma^2\psi_j^2(t_0,t)\sigma^2 \\
    % pulling the summation terms out of the square
    &\leq 4\mu_1\mu_2 BN\gamma^2\psi_j(t_0,t)  \mathbb{E}_{\cdot|\mathcal{Q}} \Bigg[ \sum_{j\neq i} q_t^{j\rightarrow i} \int_{\tau} \sum_{n\neq j} q_\tau^{n\rightarrow j} \sum_{b=0}^{B-1} \Big\|\nabla f_n(\mathbf{y}_{\tau,b}^{(n)}) - \nabla f_n(\mathbf{x}_\tau^{(n)}) \Big\|^2 \dd{\tau} \Bigg] \\
        &\quad + 4\mu_1\mu_2 B^2\gamma^2 \mathbb{E}_{\cdot|\mathcal{Q}} \Big[ \sum_{j\neq i} q_t^{j\rightarrow i} \Big\| \psi_j(t_0, t) \sum_{n\neq j} q_{\star}^{n\rightarrow j} \big[\nabla f_n(\mathbf{x}_{\star}^{(n)}) - \nabla f_j(\mathbf{x}_{\star}^{(j)}) \big]  \Big\|^2\Big]\\
        &\quad + 4\mu_1\mu_2 B^2\gamma^2 \mathbb{E}_{\cdot|\mathcal{Q}} \Bigg[ \sum_{j\neq i} q_t^{j\rightarrow i} \Bigg\| \int_{\tau} \sum_{n\neq j} q_\tau^{n\rightarrow j} \Big[\nabla f_j(\mathbf{x}_\tau^{(j)}) - \nabla f_j(\mathbf{x}_{t_0}^{(j)}) \Big] \dd{\tau} \Bigg\|^2\Bigg]\\
        &\quad + 4\mu_1\mu_2 \mathbb{E}_{\cdot|\mathcal{Q}} \Bigg[ \sum_{j\neq i} q_t^{j\rightarrow i}  \Big\|\nabla f_j(\mathbf{x}_{t_0}^{(j)}) - \nabla f_i(\mathbf{x}_{t_0}^{(i)}) \Big\|^2 \cdot \Bigg\|\gamma\int_{\tau} \sum_{n\neq j} q_\tau^{n\rightarrow j} \sum_{b=0}^{B-1} 1 \dd{\tau} \Bigg\|^2\Bigg]\\
        &\quad + \mu_1\Big(1+\frac{1}{\mu_2-1}\Big) B^2\gamma^2\psi_j^2(t_0,t) \mathbb{E}_{\cdot|\mathcal{Q}} \big[\big\| \nabla f_i(\mathbf{x}_{t_0}^{(i)}) \big\|^2\big] +\Big(1+\frac{1}{\mu_1-1}\Big) B^2N^2\gamma^2\psi_j^2(t_0,t)\sigma^2 \\
    % (d)
    % L-smoothness on the 1st and 3rd term
    % Lemma 1 on the 2nd term
    &\overset{(d)}{\leq} 4\mu_1\mu_2 BL^2N\gamma^2\psi_j(t_0,t)  \mathbb{E}_{\cdot|\mathcal{Q}} \Bigg[ \sum_{j\neq i} q_t^{j\rightarrow i} \int_{\tau} \sum_{n\neq j} q_\tau^{n\rightarrow j} \sum_{b=0}^{B-1} \big\|\mathbf{y}_{\tau,b}^{(n)} - \mathbf{x}_\tau^{(n)} \big\|^2 \dd{\tau} \Bigg]\\
        &\quad + 4\mu_1\mu_2 B^2\gamma^2\psi_j^2(t_0,t) \cdot \frac{2N\zeta^2}{N-4}\\
        &\quad + 4\mu_1\mu_2 B^2L^2\gamma^2 \mathbb{E}_{\cdot|\mathcal{Q}} \Bigg[ \sum_{j\neq i} q_t^{j\rightarrow i} \Bigg\| \int_{\tau} \sum_{n\neq j} q_\tau^{n\rightarrow j} \Big[\mathbf{x}_\tau^{(j)} - \mathbf{x}_{t_0}^{(j)} \Big] \dd{\tau} \Bigg\|^2\Bigg]\\
        &\quad+ \frac{8\mu_1\mu_2 B^2N\gamma^2\psi_j^2(t_0,t) \zeta^2}{N-4} + \mu_1\Big(1+\frac{1}{\mu_2-1}\Big) B^2\gamma^2\psi_j^2(t_0,t) \mathbb{E}_{\cdot|\mathcal{Q}} \big[\big\| \nabla f_i(\mathbf{x}_{t_0}^{(i)}) \big\|^2\big]\\
        &\quad +\Big(1+\frac{1}{\mu_1-1}\Big) B^2N^2\gamma^2\psi_j^2(t_0,t)\sigma^2 \\
    % (e)
    % 2nd term - use Lemma 1
    % 3rd term - \tau_max
    % collect the terms in the same order (... (N-4))
    &\overset{(e)}{\leq} 4\mu_1\mu_2 BL^2N\gamma^2\psi_j(t_0,t)  \mathbb{E}_{\cdot|\mathcal{Q}} \Bigg[ \sum_{j\neq i} q_t^{j\rightarrow i} \int_{\tau} \sum_{n\neq j} q_\tau^{n\rightarrow j} \sum_{b=0}^{B-1} \big\|\mathbf{y}_{\tau,b}^{(n)} - \mathbf{x}_\tau^{(n)} \big\|^2 \dd{\tau} \Bigg]\\
        &\quad + 4\mu_1\mu_2 B^2L^2\gamma^2\psi_j^2(t_0,t) \mathbb{E}_{\cdot|\mathcal{Q}} \Big[ \sum_{j\neq i} q_t^{j\rightarrow i} \big\| \mathbf{x}_{\tau_\text{max}}^{(j)} - \mathbf{x}_{t_0}^{(j)} \big\|^2\Big]\\
        &\quad+ \frac{16\mu_1\mu_2 B^2N\gamma^2\psi_j^2(t_0,t) \zeta^2}{N-4} + \mu_1\Big(1+\frac{1}{\mu_2-1}\Big) B^2\gamma^2\psi_j^2(t_0,t) \mathbb{E}_{\cdot|\mathcal{Q}} \big[\big\| \nabla f_i(\mathbf{x}_{t_0}^{(i)}) \big\|^2\big]\\
        &\quad  +\Big(1+\frac{1}{\mu_1-1}\Big) B^2N^2\gamma^2\psi_j^2(t_0,t)\sigma^2 \ ,
    \label{eq:thm_2.2.0}
\end{alignb}}
where two coefficients larger than one, denoted by $\mu_1$ and $\mu_2$, are introduced in (i) and (ii), respectively. In inequality \ref{eq:thm_2.2.0}, (a) uses \[\big\|\int_{\tau=t_0}^t \sum_{n\neq j}q_\tau^{nj} \mathbf{z}_\tau \dd{\tau}\big\|^2 \leq \psi_j(t_0, t) \int_{\tau=t_0}^t \| \sum_{n\neq j}q_\tau^{nj} \mathbf{z}_\tau \|^2\dd{\tau}\] and \[ \big\|\sum_{m=1}^M \mathbf{z}_m\big\|^2 \leq M\sum_{m=1}^M \|\mathbf{z}_m\|^2\]  for any vector $\mathbf{z}_{\star}\in\mathbb{R}^d$.\footnote{This results in $\big\|\sum_{j\in\mathcal{U}}q_{\star}^{j\rightarrow i}\mathbf{z}_j\big\|^2 \leq N\sum_{j\in\mathcal{U}} q_{\star}^{j\rightarrow i} \| \mathbf{z}_j\|^2$ and $\|\sum_{b=0}^{B-1}\mathbf{z}_b\|^2 \leq B\sum_{b=0}^{B-1}\|\mathbf{z}_b\|^2$.} (b) comes from the definition of $\sigma^2$ in Assumption \ref{assmp:sg}. In (c), Jensen's inequality is applied once again. (d) takes $L$-smoothness on the first and the second term, while Lemma \ref{lemma:dev_local_grads} is applied on the third term. In (e) the third term of inequality \ref{eq:thm_2.2.0} already contains the current lemma. Here, we introduce an index of the instant $\tau_\text{max}\in[t_0, t)$ that satisfies $\tau_\text{max}=\argmax_\tau \|\mathbf{x}_\tau^{(j)}-\mathbf{x}_{t_0}^{(j)}\|^2$.

The first term of inequality \ref{eq:thm_2.2.0}, which includes Lemma \ref{prop:ytbj_xtj}, can be rephrased as follows:
\allowdisplaybreaks{\begin{alignb}
    &\mathbb{E}_{\cdot|\mathcal{Q}} \Bigg[ \sum_{j\neq i} q_t^{j\rightarrow i} \int_{\tau} \sum_{n\neq j} q_\tau^{n\rightarrow j} \sum_{b=0}^{B-1} \big\|\mathbf{y}_{\tau,b}^{(n)} - \mathbf{x}_\tau^{(n)} \big\|^2 \dd{\tau} \Bigg]\\
    &\leq B \mathbb{E}_{\cdot|\mathcal{Q}} \Bigg[ \sum_{j\neq i} q_t^{j\rightarrow i} \int_{\tau} \sum_{n\neq j} q_\tau^{n\rightarrow j} \big\|\mathbf{y}_{\tau,\star}^{(n)} - \mathbf{x}_\tau^{(n)} \big\|^2 \dd{\tau} \Bigg] \\
    &\leq B \mathbb{E}_{\cdot|\mathcal{Q}} \Big[ \sum_{j\neq i} q_t^{j\rightarrow i} \psi_j(t_0, t) \sum_{n\neq j} q_{\star}^{n\rightarrow j} \big\|\mathbf{y}_{\star,\star}^{(n)} - \mathbf{x}_{\star}^{(n)} \big\|^2 \Big] \\
    &\leq B\psi_j(t_0, t) \mathbb{E}_{\cdot|\mathcal{Q}} \Big[ \sum_{n\neq j} q_\tau^{n\rightarrow j} \| \mathbf{y}_{\tau,b}^{(n)}-\mathbf{x}_\tau^{(n)}\|^2 \Big]
    %&:= B \mathbb{E}_{\cdot|\mathcal{Q}} \Big[ \sum_{j\neq i} q_t^{j\rightarrow i} \psi_j(t_0, t)  \big\|\mathbf{y}_{t,b}^{(j)} - \mathbf{x}_t^{(j)} \big\|^2 \Big]
    \label{eq:thm_2.2.1}
\end{alignb}}

%Plugging inequality \ref{eq:thm_2.2.1} and \ref{eq:thm_2.2.2},
We continue rephrasing the primary inequality \ref{eq:thm_2.2.0}:
\allowdisplaybreaks{\begin{align*}
    &\mathbb{E}_{\cdot|\mathcal{Q}} \Big[ \sum_{j\neq i} q_t^{j\rightarrow i} \big\| \mathbf{x}_t^{(j)} -\mathbf{x}_{t_0}^{(j)} \big\|^2\Big] \\
    % plug the two inequalitiees above
    &\leq 4\mu_1\mu_2 B^2L^2N\gamma^2\psi_j^2(t_0,t) \mathbb{E}_{\cdot|\mathcal{Q}} \Big[ \sum_{n\neq j} q_\tau^{n\rightarrow j} \| \mathbf{y}_{\tau,b}^{(n)}-\mathbf{x}_\tau^{(n)}\|^2 \Big]\\
        &\quad + 4\mu_1\mu_2 B^2L^2\gamma^2\psi_j^2(t_0,t) \mathbb{E}_{\cdot|\mathcal{Q}} \Big[ \sum_{j\neq i} q_t^{j\rightarrow i} \big\| \mathbf{x}_{\tau_\text{max}}^{(j)} - \mathbf{x}_{t_0}^{(j)} \big\|^2\Big]\\
        &\quad+ \frac{16\mu_1\mu_2 B^2N\gamma^2\psi_j^2(t_0,t) \zeta^2}{N-4} + \mu_1\Big(1+\frac{1}{\mu_2-1}\Big) B^2\gamma^2\psi_j^2(t_0,t) \mathbb{E}_{\cdot|\mathcal{Q}} \big[\big\| \nabla f_i(\mathbf{x}_{t_0}^{(i)}) \big\|^2\big]\\
        &\quad  +\Big(1+\frac{1}{\mu_1-1}\Big) B^2N^2\gamma^2\psi_j^2(t_0,t)\sigma^2
\end{align*}}
After rearranging the inequality in order to integrate those terms including $\mathbb{E}_{\cdot|\mathcal{Q}} \Big[ \sum_{j\neq i} q_\star^{j\rightarrow i} \big\| \mathbf{x}_\star^{(j)} -\mathbf{x}_{t_0}^{(j)} \big\|^2\Big]$, we have
\allowdisplaybreaks{\begin{align*}
    &\mathbb{E}_{\cdot|\mathcal{Q}} \Big[ \sum_{j\neq i} q_t^{j\rightarrow i} \big\| \mathbf{x}_t^{(j)} -\mathbf{x}_{t_0}^{(j)} \big\|^2\Big]- 4\mu_1\mu_2 B^2L^2\gamma^2\psi_j^2(t_0,t) \mathbb{E}_{\cdot|\mathcal{Q}} \Big[ \sum_{j\neq i} q_t^{j\rightarrow i} \big\| \mathbf{x}_{\tau_\text{max}}^{(j)} - \mathbf{x}_{t_0}^{(j)} \big\|^2\Big] \\
    &\leq (1-4\mu_1\mu_2 B^2L^2\gamma^2\Psi^2) \mathbb{E}_{\cdot|\mathcal{Q}} \Big[ \sum_{j\neq i} q_t^{j\rightarrow i} \big\| \mathbf{x}_t^{(j)} -\mathbf{x}_{t_0}^{(j)} \big\|^2\Big] \\
    &\leq (1-4\mu_1\mu_2 B^2L^2N\gamma^2\Psi^2) \mathbb{E}_{\cdot|\mathcal{Q}} \Big[ \sum_{j\neq i} q_t^{j\rightarrow i} \big\| \mathbf{x}_t^{(j)} -\mathbf{x}_{t_0}^{(j)} \big\|^2\Big] .
\end{align*}}
Hence, we can rephrase the inequality as
\allowdisplaybreaks{\begin{alignb}
    &\mathbb{E}_{\cdot|\mathcal{Q}} \Big[ \sum_{j\neq i} q_t^{j\rightarrow i} \big\| \mathbf{x}_t^{(j)} -\mathbf{x}_{t_0}^{(j)} \big\|^2\Big] \\
    % plug the two inequalitiees above
    &\leq \frac{1}{1-4\mu_1\mu_2 B^2L^2N\gamma^2\Psi^2} \cdot \Bigg[ 4\mu_1\mu_2 B^2L^2N\gamma^2\Psi^2 \mathbb{E}_{\cdot|\mathcal{Q}} \Big[ \sum_{n\neq j} q_\tau^{n\rightarrow j} \| \mathbf{y}_{\tau,b}^{(n)}-\mathbf{x}_\tau^{(n)}\|^2 \Big]\\
        &\quad+ \frac{16\mu_1\mu_2 B^2N\gamma^2\zeta^2\Psi^2}{N-4} + \mu_1\Big(1+\frac{1}{\mu_2-1}\Big) B^2\gamma^2\Psi^2 \mathbb{E}_{\cdot|\mathcal{Q}} \big[\big\| \nabla f_i(\mathbf{x}_{t_0}^{(i)}) \big\|^2\big]\\
        &\quad +\Big(1+\frac{1}{\mu_1-1}\Big) B^2N^2\gamma^2\sigma^2\Psi^2 \Bigg]\\
    % \mu_1 = 1/(6BLN\gamma\Psi) , c = 1/(BL\gamma\Psi)
    &\overset{(i)}{\leq} 3 \cdot \Bigg[ \frac{2}{3} \mathbb{E}_{\cdot|\mathcal{Q}} \Big[ \sum_{n\neq j} q_\tau^{n\rightarrow j} \| \mathbf{y}_{\tau,b}^{(n)}-\mathbf{x}_\tau^{(n)}\|^2 \Big]+ \frac{8\zeta^2}{3L^2(N-4)}  + \frac{B\gamma\Psi}{6LN(1-BL\gamma\Psi)} \mathbb{E}_{\cdot|\mathcal{Q}} \big[\big\| \nabla f_i(\mathbf{x}_{t_0}^{(i)}) \big\|^2\big] \\
        &\qquad +\frac{B^2N^2\gamma^2\sigma^2\Psi^2}{1-6BLN\gamma\Psi} \Bigg]\\
    &= 2 \mathbb{E}_{\cdot|\mathcal{Q}} \Big[ \sum_{n\neq j} q_\tau^{n\rightarrow j} \| \mathbf{y}_{\tau,b}^{(n)}-\mathbf{x}_\tau^{(n)}\|^2 \Big]+ \frac{8\zeta^2}{L^2(N-4)}  + \frac{B\gamma\Psi}{2LN(1-BL\gamma\Psi)} \mathbb{E}_{\cdot|\mathcal{Q}} \big[\big\| \nabla f_i(\mathbf{x}_{t_0}^{(i)}) \big\|^2\big] \\
        &\quad +\frac{3B^2N^2\gamma^2\sigma^2\Psi^2}{1-6BLN\gamma\Psi}\ ,
    % -----------------------------------
    %\\&\leq 3\cdot \Bigg[ \frac{2}{3} \mathbb{E}_{\cdot|\mathcal{Q}} \Big[ \sum_{j\neq i} q_t^{j\rightarrow i} \big\|\mathbf{y}_{t,b}^{(j)} - \mathbf{x}_t^{(j)} \big\|^2 \Big] +  \frac{4\zeta^2}{3L^2(N-4)}\\
    %    &\quad +  \frac{2}{9L^2N(1-3\sqrt{2}BLNn_t\gamma)} \mathbb{E}_{\cdot|\mathcal{Q}} \big[\big\| \nabla f_i(\mathbf{x}_{t_0}^{(i)}) \big\|^2\big] +\frac{4B^2N^2n_t^2\gamma^2\sigma^2}{4-3\sqrt{2}BLn_t\gamma} \Bigg] \\
    %&\leq 2\mathbb{E}_{\cdot|\mathcal{Q}} \Big[ \sum_{j\neq i} q_t^{j\rightarrow i} \big\|\mathbf{y}_{t,b}^{(j)} - \mathbf{x}_t^{(j)} \big\|^2 \Big] + \frac{4\zeta^2}{L^2(N-4)}\\
    %    &\quad +  \frac{2}{3L^2N(1-3\sqrt{2}BLNn_t\gamma)} \mathbb{E}_{\cdot|\mathcal{Q}} \big[\big\| \nabla f_i(\mathbf{x}_{t_0}^{(i)}) \big\|^2\big] +\frac{4B^2N^2n_t^2\gamma^2\sigma^2}{3(4-3\sqrt{2}BLn_t\gamma)}\ .
\label{eq:thm_2.2.3}
\end{alignb}}
where (i) $\mu_1=\frac{1}{6BLN\gamma\Psi}$ and $c=\frac{1}{BL\gamma\Psi}$ are applied.
Additionally, if $\Psi>0$, $\gamma\leq\frac{1}{8BLN\Psi}$ is the tighter upper bound than $\gamma\leq\frac{1}{8BL}$. With this remark, the upper bound in inequality \ref{eq:thm_2.2.3} can be simplified even more as
\allowdisplaybreaks{\begin{alignb}
    &\mathbb{E}_{\cdot|\mathcal{Q}} \Big[ \sum_{j\neq i} q_t^{j\rightarrow i} \big\| \mathbf{x}_t^{(j)} -\mathbf{x}_{t_0}^{(j)} \big\|^2\Big] \\
    &\leq 2 \mathbb{E}_{\cdot|\mathcal{Q}} \Big[ \sum_{n\neq j} q_\tau^{n\rightarrow j} \| \mathbf{y}_{\tau,b}^{(n)}-\mathbf{x}_\tau^{(n)}\|^2 \Big]+ \frac{8\zeta^2}{L^2(N-4)} \\
        &\quad + \frac{\frac{1}{8LN}}{2LN(1-\frac{1}{8N})} \mathbb{E}_{\cdot|\mathcal{Q}} \big[\big\| \nabla f_i(\mathbf{x}_{t_0}^{(i)}) \big\|^2\big]
        +\frac{3B^2N^2\sigma^2\Psi^2\cdot\frac{1}{64B^2L^2N^2\Psi^2}}{1-\frac{6BLN\Psi}{8BLN\Psi}} \\
    &\leq 2 \mathbb{E}_{\cdot|\mathcal{Q}} \Big[ \sum_{n\neq j} q_\tau^{n\rightarrow j} \| \mathbf{y}_{\tau,b}^{(n)}-\mathbf{x}_\tau^{(n)}\|^2 \Big]+ \frac{8\zeta^2}{L^2(N-4)} \\
        &\quad + \frac{1}{2L^2N(8N-1)} \mathbb{E}_{\cdot|\mathcal{Q}} \big[\big\| \nabla f_i(\mathbf{x}_{t_0}^{(i)}) \big\|^2\big]
        +\frac{3\sigma^2}{16L^2} \\
    &\leq 2 \mathbb{E}_{\cdot|\mathcal{Q}} \Big[ \sum_{n\neq j} q_\tau^{n\rightarrow j} \| \mathbf{y}_{\tau,b}^{(n)}-\mathbf{x}_\tau^{(n)}\|^2 \Big]+ \frac{8\zeta^2}{L^2(N-4)} \\
        &\quad + \frac{1}{16L^2N^2} \mathbb{E}_{\cdot|\mathcal{Q}} \big[\big\| \nabla f_i(\mathbf{x}_{t_0}^{(i)}) \big\|^2\big]
        +\frac{3\sigma^2}{16L^2}\\
    &\overset{(a)}{\leq} 2 \mathbb{E}_{\cdot|\mathcal{Q}} \Big[ \sum_{j\neq i} q_t^{j\rightarrow i} \| \mathbf{y}_{\tau,b}^{(j)}-\mathbf{x}_\tau^{(j)}\|^2 \Big]+ \frac{8\zeta^2}{L^2(N-4)} \\
        &\quad + \frac{1}{16L^2N^2} \mathbb{E}_{\cdot|\mathcal{Q}} \big[\big\| \nabla f_i(\mathbf{x}_{t_0}^{(i)}) \big\|^2\big]
        +\frac{3\sigma^2}{16L^2}\ ,
    %-------------------------------------
    %&\leq 2\mathbb{E}_{\cdot|\mathcal{Q}} \Big[ \sum_{j\neq i} q_t^{j\rightarrow i} \big\|\mathbf{y}_{t,b}^{(j)} - \mathbf{x}_t^{(j)} \big\|^2 \Big] + \frac{4\zeta^2}{L^2(N-4)}\\
    %    &\quad +  \frac{1}{L^2N} \mathbb{E}_{\cdot|\mathcal{Q}} \big[\big\| \nabla f_i(\mathbf{x}_{t_0}^{(i)}) \big\|^2\big] +\frac{2\sigma^2}{273L^2(4-\frac{1}{3N})} \\
    %&\overset{(a)}{\leq} 2\mathbb{E}_{\cdot|\mathcal{Q}} \Big[ \sum_{j\neq i} q_t^{j\rightarrow i} \big\|\mathbf{y}_{t,b}^{(j)} - \mathbf{x}_t^{(j)} \big\|^2 \Big] + \frac{4\zeta^2}{L^2(N-4)}\\
    %    &\quad +  \frac{1}{L^2N} \mathbb{E}_{\cdot|\mathcal{Q}} \big[\big\| \nabla f_i(\mathbf{x}_{t_0}^{(i)}) \big\|^2\big] +\frac{2\sigma^2}{273L^2\cdot\frac{47}{12}} \\
    %&\overset{(b)}{\leq} 2\mathbb{E}_{\cdot|\mathcal{Q}} \Big[ \sum_{j\neq i} q_t^{j\rightarrow i} \big\|\mathbf{y}_{t,b}^{(j)} - \mathbf{x}_t^{(j)} \big\|^2 \Big] + \frac{4\zeta^2}{L^2(N-4)}\\
    %    &\quad +  \frac{1}{L^2N} \mathbb{E}_{\cdot|\mathcal{Q}} \big[\big\| \nabla f_i(\mathbf{x}_{t_0}^{(i)}) \big\|^2\big] +\frac{\sigma^2}{500L^2}\ ,
\label{eq:thm_2.2}
\end{alignb}}
where (a) is satisfied without loss of generality. %where (a) uses $N\geq4$, and (b) is because of $\frac{24}{273\cdot47}\leq\frac{1}{500}$.
\end{proof}
\noindent\textbf{Remark.}\quad This proposition appears in Lemma \ref{lemma:xt0j_xtj}, which is then used at the second term of inequality 
 \ref{eq:thm_2}.

%=====================================================
%\pagebreak
\subsection{Lemmas}\label{appendix:lemmas}
In collaborative learning, local computations often occur more frequently than communication. This is to avoid duplicating transmissions, which can occur in the reverse scenario.

\begin{lemma}{} \label{lemma:floor_t_and_t}
    For all $n\in\mathcal{U}$, there are no fewer $\mathbf{y}_{t, b}^{(n)}$ than $\mathbf{y}_{\lfloor t \rfloor, b}^{(n)}$ within any given range of time $\{t | t\in [t_0, t_0+P)\}$. In other words, it also satisfies that
    \begin{align}
        \Big\| \int_{P} \sum_{b=0}^{B-1} \mathbf{g}_n(\mathbf{y}_{\lfloor t \rfloor, b}^{(n)}) \dd{t} \Big\|^2 \leq \Big\| \int_{P} \sum_{b=0}^{B-1} \mathbf{g}_n(\mathbf{y}_{t, b}^{(n)}) \dd{t}\Big\|^2.
    \end{align}
\end{lemma}
\begin{figure}[t]
    \centering
    \includegraphics[width=0.55\columnwidth]{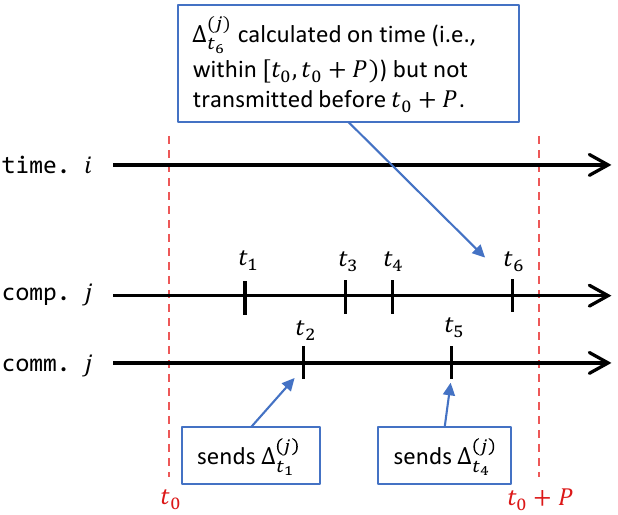}
    \caption{The latest local update of $j$ is unable to be transmitted within the given range $[t_0, t_0+P)$ because of the independence between computation timestamps and communication (transmission) timestamps.}
    \label{fig:issue_within_tau}
\end{figure}

\begin{proof}
    \normalsize In Fig. (\ref{fig:issue_within_tau}), the value of $\Delta_{t_4}^{(j)}$ can differ from $\Delta_{t_3}^{(j)}$ if another node transmits a message to node $j$, thereby affecting the value of $\mathbf{x}^{(j)}$. To facilitate our analysis, we assume that each user creates a backup of the non transmitted local updates for the upcoming transmission event. Returning to the scenario depicted in Fig. (\ref{fig:issue_within_tau}), based on this assumption, user $j$ sends both $\Delta_{t_3}^{(j)}$ and $\Delta_{t_4}^{(j)}$ to user $i$ at the earliest transmission event time, which is $t_5$.     
\end{proof}

%\subsubsection{Lem A2}
\begin{lemma}{(Upper bound for superpositioned model deviations.)} \label{lemma:ytbj_xtj}
    For all $i,j\in\mathcal{U}$, when $\gamma\leq\min(\frac{1}{8BL},\frac{1}{8BLN\Psi})$, we have
    \begin{align*}
        &\sum_{j\neq i} q_t^{j\rightarrow i} \mathbb{E}_{\cdot|\mathcal{Q}}\big[ \| \mathbf{y}_{t,b}^{(j)} - \mathbf{x}_t^{(j)}\|^2 \big] \\
        &\leq \frac{16\zeta^2}{L^2(N-4)} +\frac{3\sigma^2}{8L^2} +\frac{9N\zeta^2}{2L^2} +16B\gamma^2\sigma^2 +\Big(\frac{1}{8L^2N}+128B^2\gamma^2 \Big) \mathbb{E}_{\cdot|\mathcal{Q}} \big[\big\| \nabla f_i(\mathbf{x}_{t_0}^{(i)}) \big\|^2\big] \ .
    \end{align*}
\end{lemma}

\begin{proof}
    Proposition \ref{prop:ytbj_xtj} and Proposition \ref{prop:xt0j_xtj} can be interpreted as a system of linear inequalities. %Used Proposition \ref{prop:ytbj_xtj}.
    Applying (a) Proposition \ref{prop:ytbj_xtj} and (b) Proposition \ref{prop:xt0j_xtj} respectively, we get
    \allowdisplaybreaks{\begin{align*}
        &\sum_{j\neq i} q_t^{j\rightarrow i} \mathbb{E}_{\cdot|\mathcal{Q}}\big[ \| \mathbf{y}_{t,b}^{(j)} - \mathbf{x}_t^{(j)}\|^2 \big] \\
        &\overset{(a)}{\leq} \frac{2}{5}\sum_{j\neq i}q_t^{j\rightarrow i}\mathbb{E}_{\cdot|\mathcal{Q}} \big[\big\| \mathbf{x}_t^{(i)}-\mathbf{x}_{t_0}^{(i)} \big\|^2\big] +\frac{9N\zeta^2}{10L^2} +\frac{16B\gamma^2\sigma^2}{5} +\frac{128B^2\gamma^2}{5}\mathbb{E}_{\cdot|\mathcal{Q}}[\| \nabla f_i(\mathbf{x}_{t_0}^{(i)})\|^2] \\
        &\overset{(b)}{\leq} \frac{2}{5} \Big( 2 \mathbb{E}_{\cdot|\mathcal{Q}} \Big[ \sum_{j\neq i} q_t^{j\rightarrow i} \| \mathbf{y}_{t,b}^{(j)}-\mathbf{x}_t^{(j)}\|^2 \Big]+ \frac{8\zeta^2}{L^2(N-4)} + \frac{1}{16L^2N^2} \mathbb{E}_{\cdot|\mathcal{Q}} \big[\big\| \nabla f_i(\mathbf{x}_{t_0}^{(i)}) \big\|^2\big]     +\frac{3\sigma^2}{16L^2}\Big)\\
            &\quad+\frac{9N\zeta^2}{10L^2} +\frac{16B\gamma^2\sigma^2}{5} +\frac{128B^2\gamma^2}{5}\mathbb{E}_{\cdot|\mathcal{Q}}[\| \nabla f_i(\mathbf{x}_{t_0}^{(i)})\|^2] \\
        &= \frac{4}{5} \mathbb{E}_{\cdot|\mathcal{Q}} \Big[ \sum_{j\neq i} q_t^{j\rightarrow i} \big\|\mathbf{y}_{t,b}^{(j)} - \mathbf{x}_t^{(j)} \big\|^2 \Big] + \frac{16\zeta^2}{5L^2(N-4)} +\frac{1}{40L^2N} \mathbb{E}_{\cdot|\mathcal{Q}} \big[\big\| \nabla f_i(\mathbf{x}_{t_0}^{(i)}) \big\|^2\big] +\frac{3\sigma^2}{40L^2}\\
            &\quad +\frac{9N\zeta^2}{10L^2} +\frac{16B\gamma^2\sigma^2}{5} +\frac{128B^2\gamma^2}{5}\mathbb{E}_{\cdot|\mathcal{Q}} \big[\big\| \nabla f_i(\mathbf{x}_{t_0}^{(i)}) \big\|^2\big] \ ,
    \end{align*}}
    and therefore,
    \allowdisplaybreaks{\begin{align*}
        &\frac{1}{5}\sum_{j\neq i} q_t^{j\rightarrow i} \mathbb{E}_{\cdot|\mathcal{Q}}\big[ \| \mathbf{y}_{t,b}^{(j)} - \mathbf{x}_t^{(j)}\|^2 \big] \\
        &\leq \frac{16\zeta^2}{5L^2(N-4)} +\frac{3\sigma^2}{40L^2} +\frac{9N\zeta^2}{10L^2} +\frac{16B\gamma^2\sigma^2}{5} +\Big(\frac{1}{40L^2N}+\frac{128B^2\gamma^2}{5} \Big) \mathbb{E}_{\cdot|\mathcal{Q}} \big[\big\| \nabla f_i(\mathbf{x}_{t_0}^{(i)}) \big\|^2\big]\ .
    \end{align*}}
\end{proof}

%\subsubsection{Lem A3}
\begin{lemma}{(Upper bound for the local reference model change)} \label{lemma:xt0j_xtj} %\tgreen{[This lemma appears at the second term of inequality  \ref{eq:thm_2}.]}
    When Assumption \ref{assmp:finite_Psi} holds true during $[t_0, t)$ for all users (i.e., when the number of events during the given period $[t_0, t)$ is finite) and $\gamma\leq\min(\frac{1}{8BL}, \frac{1}{8BLN\Psi})$, we have
    \begin{align*}
        &\mathbb{E}_{\cdot|\mathcal{Q}} \Big[ \sum_{j\neq i} q_t^{j\rightarrow i} \big\| \mathbf{x}_t^{(j)} -\mathbf{x}_{t_0}^{(j)} \big\|^2\Big] \\
        %&\leq 9N\zeta^2 +32B\gamma^2\sigma^2 +\frac{20\zeta^2}{L^2(N-4)}  +\frac{\sigma^2}{100L^2} +\Big(32B\gamma^2+\frac{5}{L^2N}\Big)\mathbb{E}_{\cdot|\mathcal{Q}}[\| \nabla f_i(\mathbf{x}_{t_0}^{(i)})\|^2] \ . \\
        &\leq \frac{9N\zeta^2}{L^2} +32B\gamma^2\sigma^2 + \frac{40\zeta^2}{L^2(N-4)} +\frac{15\sigma^2}{16L^2} + \Big(256B^2\gamma^2+\frac{5}{16L^2N^2}\Big)\mathbb{E}_{\cdot|\mathcal{Q}} \big[\big\| \nabla f_i(\mathbf{x}_{t_0}^{(i)}) \big\|^2\big] \ .
    \end{align*}
\end{lemma}

\begin{proof}
    Approaching in the same fashion of proving as in Lemma \ref{lemma:ytbj_xtj}, we have
    \allowdisplaybreaks{\begin{align*}
        &\mathbb{E}_{\cdot|\mathcal{Q}} \Big[ \sum_{j\neq i} q_t^{j\rightarrow i} \big\| \mathbf{x}_t^{(j)} -\mathbf{x}_{t_0}^{(j)} \big\|^2\Big] \\
        &\overset{(a)}{\leq} 2 \mathbb{E}_{\cdot|\mathcal{Q}} \Big[ \sum_{j\neq i} q_t^{j\rightarrow i} \| \mathbf{y}_{t,b}^{(j)}-\mathbf{x}_t^{(j)}\|^2 \Big]+ \frac{8\zeta^2}{L^2(N-4)} + \frac{1}{16L^2N^2} \mathbb{E}_{\cdot|\mathcal{Q}} \big[\big\| \nabla f_i(\mathbf{x}_{t_0}^{(i)}) \big\|^2\big]     +\frac{3\sigma^2}{16L^2} \\
        &\overset{(b)}{\leq} 2 \Big( \frac{2}{5}\sum_{j\neq i}q_t^{j\rightarrow i}\mathbb{E}_{\cdot|\mathcal{Q}} \big[\big\| \mathbf{x}_t^{(i)}-\mathbf{x}_{t_0}^{(i)} \big\|^2\big]  +\frac{9N\zeta^2}{10L^2} +\frac{16B\gamma^2\sigma^2}{5} +\frac{128B^2\gamma^2}{5}\mathbb{E}_{\cdot|\mathcal{Q}}[\| \nabla f_i(\mathbf{x}_{t_0}^{(i)})\|^2] \Big)\\
            &\quad + \frac{8\zeta^2}{L^2(N-4)} + \frac{1}{16L^2N^2} \mathbb{E}_{\cdot|\mathcal{Q}} \big[\big\| \nabla f_i(\mathbf{x}_{t_0}^{(i)}) \big\|^2\big] +\frac{3\sigma^2}{16L^2} \\
        %&= \frac{4}{5}\sum_{j\neq i}q_t^{j\rightarrow i}\mathbb{E}_{\cdot|\mathcal{Q}} \big[\big\| \mathbf{x}_t^{(i)}-\mathbf{x}_{t_0}^{(i)} \big\|^2\big]  +\frac{9N\zeta^2}{5L^2} +\frac{32B\gamma^2\sigma^2}{5} +\frac{256B^2\gamma^2}{5}\mathbb{E}_{\cdot|\mathcal{Q}}[\| \nabla f_i(\mathbf{x}_{t_0}^{(i)})\|^2] \\
        %    &\quad + \frac{8\zeta^2}{L^2(N-4)} + \frac{1}{16L^2N^2} \mathbb{E}_{\cdot|\mathcal{Q}} \big[\big\| \nabla f_i(\mathbf{x}_{t_0}^{(i)}) \big\|^2\big] +\frac{3\sigma^2}{16L^2} \\
        &= \frac{4}{5}\sum_{j\neq i}q_t^{j\rightarrow i}\mathbb{E}_{\cdot|\mathcal{Q}} \big[\big\| \mathbf{x}_t^{(i)}-\mathbf{x}_{t_0}^{(i)} \big\|^2\big]  +\frac{9N\zeta^2}{5L^2} +\frac{32B\gamma^2\sigma^2}{5} + \frac{8\zeta^2}{L^2(N-4)} +\frac{3\sigma^2}{16L^2}\\
            &\quad + \Big(\frac{256B^2\gamma^2}{5}+\frac{1}{16L^2N^2}\Big)\mathbb{E}_{\cdot|\mathcal{Q}} \big[\big\| \nabla f_i(\mathbf{x}_{t_0}^{(i)}) \big\|^2\big] \ ,
        %&=9N\zeta^2 +32B\gamma^2\sigma^2 +\frac{20\zeta^2}{L^2(N-4)}  +\frac{\sigma^2}{100L^2} +\Big(32B\gamma^2+\frac{5}{L^2N}\Big)\mathbb{E}_{\cdot|\mathcal{Q}}[\| \nabla f_i(\mathbf{x}_{t_0}^{(i)})\|^2]\ ,
    \end{align*}}
    where (a) uses Proposition \ref{prop:xt0j_xtj}, and (b) comes from Lemma \ref{lemma:ytbj_xtj}.
\end{proof}

%==========================================================
\subsection{Proof of Theorem \ref{thm:main}}\label{appendix:proof}
The proof of Theorem \ref{thm:main} is based on the proof provided in \cite{wang20-neurips}.

Beginning with rephrasing the $L$-smoothness between $f_i(\mathbf{x}_{t_0+P}^{(i)})$ and $f_i(\mathbf{x}_{t_0}^{(i)})$, we have
{\allowdisplaybreaks
\begin{align*}
    &\mathbb{E}_{\cdot|\mathcal{Q},t_0} [f_i(\mathbf{x}_{t_0+P}^{(i)})] \\
    &\leq f_i(\mathbf{x}_{t_0}^{(i)}) + \mathbb{E}_{\cdot|\mathcal{Q},t_0} [ \bigl\langle \nabla f_i(\mathbf{x}_{t_0}^{(i)}),\ \mathbf{x}_{t_0+P}^{(i)}-\mathbf{x}_{t_0}^{(i)} \bigr\rangle ] + \frac{L}{2} \mathbb{E}_{\cdot|\mathcal{Q},t_0} \Big[ \big\|\mathbf{x}_{t_0+P}^{(i)}-\mathbf{x}_{t_0}^{(i)} \big\|^2 \Big] \\
    &\leq f_i(\mathbf{x}_{t_0}^{(i)})
    - \gamma \Bigl\langle \nabla f_i(\mathbf{x}_{t_0}^{(i)}), \mathbb{E}_{\cdot|\mathcal{Q},t_0} \Big[  \int_{P} \sum_{j\neq i} q_t^{j\rightarrow i} \sum_{b=0}^{B-1} \mathbf{g}_j(\mathbf{y}_{\lfloor t \rfloor, b}^{(j)}) \dd{t} \Big] \Bigr\rangle \\
        &\quad+\frac{\gamma^2L}{2}\mathbb{E}_{\cdot|\mathcal{Q},t_0} \Big[ \Big\| \int_{P} \sum_{j\neq i} q_t^{j\rightarrow i} \sum_{b=0}^{B-1} \mathbf{g}_j(\mathbf{y}_{\lfloor t \rfloor, b}^{(j)}) \dd{t} \Big\|^2 \Big] \\
    &= f_i(\mathbf{x}_{t_0}^{(i)})
    - \gamma \Bigl\langle \nabla f_i(\mathbf{x}_{t_0}^{(i)}), \mathbb{E}_{\cdot|\mathcal{Q},t_0} \Big[  \int_{P} \sum_{j\neq i} q_t^{j\rightarrow i} \sum_{b=0}^{B-1} \mathbb{E} \big[ \mathbf{g}_j(\mathbf{y}_{\lfloor t \rfloor, b}^{(j)}) | \mathcal{Q},\mathbf{y}_{\lfloor t\rfloor,b}^{(j)},\mathbf{x}_{t_0}^{(i)} \big] \dd{t} \Big] \Bigr\rangle \\
        &\quad+\frac{\gamma^2L}{2}\mathbb{E}_{\cdot|\mathcal{Q},t_0} \Big[ \Big\| \int_{P} \sum_{j\neq i} q_t^{j\rightarrow i} \sum_{b=0}^{B-1} \mathbf{g}_j(\mathbf{y}_{\lfloor t \rfloor, b}^{(j)}) \dd{t} \Big\|^2 \Big] \\
    &= f_i(\mathbf{x}_{t_0}^{(i)})
    - \gamma \Bigl\langle \nabla f_i(\mathbf{x}_{t_0}^{(i)}), \mathbb{E}_{\cdot|\mathcal{Q},t_0} \Big[  \int_{P} \sum_{j\neq i} q_t^{j\rightarrow i} \sum_{b=0}^{B-1} \nabla f_j (\mathbf{y}_{\lfloor t \rfloor, b}^{(j)}) \dd{t}\Big] \Bigr\rangle \\
        &\quad+\frac{\gamma^2L}{2}\mathbb{E}_{\cdot|\mathcal{Q},t_0} \Big[ \Big\| \int_{P} \sum_{j\neq i} q_t^{j\rightarrow i} \sum_{b=0}^{B-1} \mathbf{g}_j(\mathbf{y}_{\lfloor t \rfloor, b}^{(j)}) \dd{t}\Big\|^2 \Big]
\end{align*}}
Taking expectation on both sides over $\mathbf{x}_{t_0}^{(i)}$, we obtain
\begin{align}\label{proof:mainthm-1}
    \mathbb{E}_{\cdot|\mathcal{Q}}[f_i(\mathbf{x}_{t_0+P}^{(i)})] &\leq  \mathbb{E}_{\cdot|\mathcal{Q}}[f_i(\mathbf{x}_{t_0}^{(i)})] - \gamma \mathbb{E}_{\cdot|\mathcal{Q},t_0} \Big[ \Bigl\langle \nabla f_i(\mathbf{x}_{t_0}^{(i)}), \int_{P} \sum_{j\neq i} q_t^{j\rightarrow i} \sum_{b=0}^{B-1} \nabla f_j (\mathbf{y}_{\lfloor t \rfloor, b}^{(j)}) \dd{t}\Bigr\rangle \Big]  \nonumber \\
        &\quad+ \frac{\gamma^2 L}{2}  \mathbb{E}_{\cdot|\mathcal{Q}} \Big[ \Big\| \int_{P} \sum_{j\neq i} q_t^{j\rightarrow i} \sum_{b=0}^{B-1} \mathbf{g}_j(\mathbf{y}_{\lfloor t \rfloor, b}^{(j)}) \dd{t}\Big\|^2 \Big]
\end{align}

Here, we reintroduce a finite variable from Definition \ref{def:psi}, $\Psi\in\mathbb{R}^+$, to indicate the maximum total number of all message exchanging events during the time period $[t_0, t_0+ P)$. We set an assumption that $\Psi\geq 3$ for any time elapse $[t_0, t_0+P)$ in which $t_0$ is multiple to $P$.

Considering the second term in the inequality \ref{proof:mainthm-1},
{\allowdisplaybreaks
\begin{align*}
    &-\Bigl\langle \nabla f_i(\mathbf{x}_{t_0}^{(i)}), \int_{P} \sum_{j\neq i} q_t^{j\rightarrow i} \sum_{b=0}^{B-1} \nabla f_j (\mathbf{y}_{\lfloor t \rfloor, b}^{(j)}) \dd{t}\Bigr\rangle \\
    &=-\frac{1}{B\Psi}\Bigl\langle B\Psi \nabla f_i(\mathbf{x}_{t_0}^{(i)}), \int_{P} \sum_{j\neq i} q_t^{j\rightarrow i} \sum_{b=0}^{B-1} \nabla f_j (\mathbf{y}_{\lfloor t \rfloor, b}^{(j)}) \Bigr\rangle \\
    &=\frac{1}{2B\Psi}\Big\| B\Psi\nabla f_i(\mathbf{x}_{t_0}^{(i)}) - \int_{P} \sum_{j\neq i} q_t^{j\rightarrow i} \sum_{b=0}^{B-1} \nabla f_j (\mathbf{y}_{\lfloor t \rfloor, b}^{(j)}) \dd{t} \Big\|^2 \\
    &\quad- \frac{B\Psi}{2}\| \nabla f_i(\mathbf{x}_{t_0}^{(i)})\|^2 - \frac{1}{2B\Psi}\Big\| \int_{P} \sum_{j\neq i} q_t^{j\rightarrow i} \sum_{b=0}^{B-1} \nabla f_j (\mathbf{y}_{\lfloor t \rfloor, b}^{(j)})\dd{t}\Big\|^2 \\
    &=\frac{1}{2B\Psi}\Big\| \int_{P} \sum_{j\neq i} q_t^{j\rightarrow i} \sum_{b=0}^{B-1} \big[\nabla f_i(\mathbf{x}_{t_0}^{(i)})- \nabla f_j (\mathbf{y}_{\lfloor t \rfloor, b}^{(j)})\big]\dd{t} \Big\|^2 \\
    &\quad- \frac{B\Psi}{2}\| \nabla f_i(\mathbf{x}_{t_0}^{(i)})\|^2 - \frac{1}{2B\Psi}\Big\| \int_{P} \sum_{j\neq i} q_t^{j\rightarrow i} \sum_{b=0}^{B-1} \nabla f_j (\mathbf{y}_{\lfloor t \rfloor, b}^{(j)})\dd{t} \Big\|^2 \\
    &=\frac{1}{2B\Psi}\Big\| \int_{P} \sum_{j\neq i} q_t^{j\rightarrow i} \sum_{b=0}^{B-1} [\nabla f_i(\mathbf{x}_{t_0}^{(i)})- \nabla f_j (\mathbf{y}_{\lfloor t \rfloor, b}^{(j)}) ] \dd{t}\Big\|^2 \\
    &\quad- \frac{B\Psi}{2}\| \nabla f_i(\mathbf{x}_{t_0}^{(i)})\|^2 - \frac{1}{2B\Psi}\Big\| \int_{P} \sum_{j\neq i} q_t^{j\rightarrow i} \sum_{b=0}^{B-1} \nabla f_j (\mathbf{y}_{\lfloor t \rfloor, b}^{(j)})\dd{t}\Big\|^2
    \end{align*}}
In order to deal with two variables controlled by different agents, two terms are added and subtracted for further proof: local model gradient calculated by $j$ and its local reference model, respectively.
{\allowdisplaybreaks
\begin{alignb}
    % adding-subtracting terms
    %&\overset{\tred{(*)}}{=}
    &=\frac{1}{2B\Psi}\Big\| \int_{P} \sum_{j\neq i} q_t^{j\rightarrow i} \sum_{b=0}^{B-1} \Big[\nabla f_i(\mathbf{x}_{t_0}^{(i)}) -\nabla f_j(\mathbf{x}_{t_0}^{(j)}) +\nabla f_j(\mathbf{x}_{t_0}^{(j)}) -\nabla f_j\big( \mathbf{x}_t^{(j)}\big) +\nabla f_j\big(\mathbf{x}_t^{(j)}\big)  -\nabla f_j (\mathbf{y}_{\lfloor t \rfloor, b}^{(j)}) \Big] \dd{t}\Big\|^2\\
        &\quad - \frac{B\Psi}{2}\| \nabla f_i(\mathbf{x}_{t_0}^{(i)})\|^2 - \frac{1}{2B\Psi}\Big\| \int_{P} \sum_{j\neq i} q_t^{j\rightarrow i} \sum_{b=0}^{B-1} \nabla f_j (\mathbf{y}_{\lfloor t \rfloor, b}^{(j)})\dd{t}\Big\|^2 \\
    % ||a+b+c||^2 \leq 3(a^2+b^2+c^2)
    &\leq \frac{3}{2B\Psi} \Big\| \int_{P} \sum_{j\neq i} q_t^{j\rightarrow i} \sum_{b=0}^{B-1} \big[ \nabla f_j\big( \mathbf{x}_t^{(j)}\big) - \nabla f_j (\mathbf{y}_{\lfloor t \rfloor, b}^{(j)}) \big] \dd{t}\Big\|^2 \\
        &\quad +\frac{3}{2B\Psi}\Big\| \int_{P} \sum_{j\neq i} q_t^{j\rightarrow i} \sum_{b=0}^{B-1} \big[ \nabla f_j(\mathbf{x}_{t_0}^{(j)}) - \nabla f_j\big( \mathbf{x}_t^{(j)}\big) \big] \dd{t} \Big\|^2  \\
        &\quad +\frac{3}{2B\Psi}\Big\| \int_{P} \sum_{j\neq i} q_t^{j\rightarrow i} \sum_{b=0}^{B-1} [\nabla f_i(\mathbf{x}_{t_0}^{(i)}) -\nabla f_j(\mathbf{x}_{t_0}^{(j)})] \dd{t}\Big\|^2 \\
        &\quad- \frac{B\Psi}{2}\| \nabla f_i(\mathbf{x}_{t_0}^{(i)})\|^2 - \frac{1}{2B\Psi}\Big\| \int_{P} \sum_{j\neq i} q_t^{j\rightarrow i} \sum_{b=0}^{B-1} \nabla f_j (\mathbf{y}_{\lfloor t \rfloor, b}^{(j)})\dd{t} \Big\|^2 \\
    % (a) pulling the main terms out of the L2 norms
    %     applied to the first three terms
    &\overset{(a)}{\leq} \frac{3}{2B\Psi} \Big\| B\Psi \sum_{j\neq i} q_t^{j\rightarrow i} \big[ \nabla f_j\big( \mathbf{x}_t^{(j)}\big) - \nabla f_j (\mathbf{y}_{\lfloor t \rfloor, b}^{(j)}) \big] \Big\|^2 \\
        &\quad +\frac{3}{2B\Psi}\Big\| B\Psi \sum_{j\neq i} q_t^{j\rightarrow i} \big[ \nabla f_j(\mathbf{x}_{t_0}^{(j)}) - \nabla f_j\big( \mathbf{x}_t^{(j)}\big) \big] \Big\|^2  \\
        &\quad +\frac{3}{2B\Psi}\Big\| B\Psi \sum_{j\neq i} q_t^{j\rightarrow i} [\nabla f_i(\mathbf{x}_{t_0}^{(i)}) -\nabla f_j(\mathbf{x}_{t_0}^{(j)})] \Big\|^2 \\
        &\quad- \frac{B\Psi}{2}\| \nabla f_i(\mathbf{x}_{t_0}^{(i)})\|^2 - \frac{1}{2B\Psi}\Big\| \int_{P} \sum_{j\neq i} q_t^{j\rightarrow i} \sum_{b=0}^{B-1} \nabla f_j (\mathbf{y}_{\lfloor t \rfloor, b}^{(j)})\dd{t} \Big\|^2 \\
    % (term in Lemma A2, A3) <= N * (term in main thm)
    % because q^2 <= q for all q\in[0,1].
    &\overset{(b)}{\leq} \frac{3BN\Psi}{2} \sum_{j\neq i} q_t^{j\rightarrow i} \Big\|  \nabla f_j\big( \mathbf{x}_t^{(j)}\big) - \nabla f_j (\mathbf{y}_{\lfloor t \rfloor, b}^{(j)}) \Big\|^2 \\
        &\quad +\frac{3BN\Psi}{2} \sum_{j\neq i} q_t^{j\rightarrow i} \Big\| \nabla f_j(\mathbf{x}_{t_0}^{(j)}) - \nabla f_j\big( \mathbf{x}_t^{(j)}\big) \Big\|^2  \\
        &\quad +\frac{3B\Psi}{2} \Big\| \sum_{j\neq i} q_t^{j\rightarrow i} \big[\nabla f_i(\mathbf{x}_{t_0}^{(i)}) -\nabla f_j(\mathbf{x}_{t_0}^{(j)})\big] \Big\|^2 \\
        &\quad- \frac{B\Psi}{2}\| \nabla f_i(\mathbf{x}_{t_0}^{(i)})\|^2 - \frac{1}{2B\Psi}\Big\| \int_{P} \sum_{j\neq i} q_t^{j\rightarrow i} \sum_{b=0}^{B-1} \nabla f_j (\mathbf{y}_{\lfloor t \rfloor, b}^{(j)})\dd{t} \Big\|^2 \\
    % (c) Lemma 1: || int_tau sum_j q_t||^2 is smaller than 2N\zeta^2/(N-4)
    %              applied to the third term
    &\overset{(c)}{\leq} \frac{3BL^2N\Psi}{2} \sum_{j\neq i} q_t^{j\rightarrow i} \big\| \mathbf{x}_t^{(j)} - \mathbf{y}_{\lfloor t \rfloor, b}^{(j)} \big\|^2  +\frac{3BL^2N\Psi}{2} \sum_{j\neq i} q_t^{j\rightarrow i} \big\| \mathbf{x}_{t_0}^{(j)} - \mathbf{x}_t^{(j)} \big\|^2 +\frac{3BN\Psi\zeta^2}{N-4} \\
        &\quad- \frac{B\Psi}{2}\| \nabla f_i(\mathbf{x}_{t_0}^{(i)})\|^2 - \frac{1}{2B\Psi}\Big\| \int_{P} \sum_{j\neq i} q_t^{j\rightarrow i} \sum_{b=0}^{B-1} \nabla f_j (\mathbf{y}_{\lfloor t \rfloor, b}^{(j)})\dd{t} \Big\|^2
    \label{eq:thm_2.1} 
    % &\overset{(x)}{\leq} \frac{3}{2BP}\Big\| \int_{P} \sum_{j\neq i} q_t^{j\rightarrow i} \sum_{b=0}^{B-1} [\nabla f(\mathbf{x}_{t_0}^{(i)}) -\nabla f_j(\mathbf{x}_{t_0}^{(i)})]\Big\|^2 
\end{alignb}}
where (a) reflects Jensen's inequality on the first three terms, pulling the terms out of the L2 norms; (b) is valid because $\|\sum_{j=1}^N q(j)\mathbf{z}(j)\|^2 \leq N\sum_{j=1}^N q(j)\|\mathbf{z}(j)\|^2$ for all $q_\star\in[0,1]$; (c) $L$-smoothness on the first two terms and Lemma \ref{lemma:dev_local_grads} on the third term.

Hence, the expectation can be bounded as follows:
\allowdisplaybreaks{\begin{alignb}
    &\mathbb{E}_{\cdot|\mathcal{Q}} \Big[ -\Bigl\langle \nabla f_i(\mathbf{x}_{t_0}^{(i)}), \int_{P} \sum_{j\neq i} q_t^{j\rightarrow i} \sum_{b=0}^{B-1} \nabla f_j (\mathbf{y}_{\lfloor t \rfloor, b}^{(j)}) \dd{t}\Bigr\rangle \Big] \\
    &\leq \frac{3BL^2N\Psi}{2} \mathbb{E}_{\cdot|\mathcal{Q}} \Big[ \sum_{j\neq i} q_t^{j\rightarrow i} \big\| \mathbf{x}_t^{(j)} - \mathbf{y}_{\lfloor t \rfloor, b}^{(j)} \big\|^2 \Big] \\
        &\quad +\frac{3BL^2N\Psi}{2} \mathbb{E}_{\cdot|\mathcal{Q}} \Big[ \sum_{j\neq i} q_t^{j\rightarrow i} \big\| \mathbf{x}_{t_0}^{(j)} - \mathbf{x}_t^{(j)} \big\|^2\Big] +\frac{3BN\Psi\zeta^2}{N-4}  \\
        &\quad- \frac{B\Psi}{2} \mathbb{E}_{\cdot|\mathcal{Q}} [\| \nabla f_i(\mathbf{x}_{t_0}^{(i)})\|^2] - \frac{1}{2B\Psi} \mathbb{E}_{\cdot|\mathcal{Q}} \Big[ \Big\| \int_{P} \sum_{j\neq i} q_t^{j\rightarrow i} \sum_{b=0}^{B-1} \nabla f_j (\mathbf{y}_{\lfloor t \rfloor, b}^{(j)}) \dd{t}\Big\|^2 \Big] \\
    % (a) Lemma a2 and a3
    &\overset{(a)}{\leq} \frac{3BL^2N\Psi}{2} \Bigg[ \frac{16\zeta^2}{L^2(N-4)} +\frac{3\sigma^2}{8L^2} +\frac{9N\zeta^2}{2L^2} +16B\gamma^2\sigma^2 +\Big(\frac{1}{8L^2N}+128B^2\gamma^2 \Big) \mathbb{E}_{\cdot|\mathcal{Q}} \big[\big\| \nabla f_i(\mathbf{x}_{t_0}^{(i)}) \big\|^2\big] \bigg] \\
        &\quad + \frac{3BL^2N\Psi}{2} \Bigg[ \frac{9N\zeta^2}{L^2}+32B\gamma^2\sigma^2 + \frac{40\zeta^2}{L^2(N-4)} +\frac{15\sigma^2}{16L^2} + \Big(256B^2\gamma^2+\frac{5}{16L^2N^2}\Big)\mathbb{E}_{\cdot|\mathcal{Q}} \big[\big\| \nabla f_i(\mathbf{x}_{t_0}^{(i)}) \big\|^2\big] \bigg] \\
        &\quad + \frac{3BN\Psi\zeta^2}{N-4} - \frac{B\Psi}{2} \mathbb{E}_{\cdot|\mathcal{Q}} [\| \nabla f_i(\mathbf{x}_{t_0}^{(i)})\|^2] - \frac{1}{2B\Psi} \mathbb{E}_{\cdot|\mathcal{Q}} \Big[ \Big\| \int_{P} \sum_{j\neq i} q_t^{j\rightarrow i} \sum_{b=0}^{B-1} \nabla f_j (\mathbf{y}_{\lfloor t \rfloor, b}^{(j)}) \dd{t}\Big\|^2 \Big] \\
    &=\frac{3BL^2N\Psi}{2} \Bigg[ \frac{56\zeta^2}{L^2(N-4)} +\frac{21\sigma^2}{16L^2} +\frac{27N\zeta^2}{2L^2} +48B\gamma^2\sigma^2 +\Big(\frac{7}{16L^2N}+384B^2\gamma^2 \Big) \mathbb{E}_{\cdot|\mathcal{Q}} \big[\big\| \nabla f_i(\mathbf{x}_{t_0}^{(i)}) \big\|^2\big] \Bigg] \\
        &\quad + \frac{3BN\Psi\zeta^2}{N-4} - \frac{B\Psi}{2} \mathbb{E}_{\cdot|\mathcal{Q}} [\| \nabla f_i(\mathbf{x}_{t_0}^{(i)})\|^2] - \frac{1}{2B\Psi} \mathbb{E}_{\cdot|\mathcal{Q}} \Big[ \Big\| \int_{P} \sum_{j\neq i} q_t^{j\rightarrow i} \sum_{b=0}^{B-1} \nabla f_j (\mathbf{y}_{\lfloor t \rfloor, b}^{(j)}) \dd{t}\Big\|^2 \Big] \\
    % arranging the terms
    &=  \frac{87BN\zeta^2\Psi}{N-4} +\frac{63BN\sigma^2\Psi}{32} +\frac{81BN^2\zeta^2\Psi}{4} +72B^2L^2N\gamma^2\sigma^2\Psi \\
        &\quad +B\Psi\Big(\frac{21}{32N}+576B^2L^2N\gamma^2 -\frac{1}{2}\Big) \mathbb{E}_{\cdot|\mathcal{Q}} \big[\big\| \nabla f_i(\mathbf{x}_{t_0}^{(i)}) \big\|^2\big]\\
        &\quad -\frac{1}{2B\Psi} \mathbb{E}_{\cdot|\mathcal{Q}} \Big[ \Big\| \int_{P} \sum_{j\neq i} q_t^{j\rightarrow i} \sum_{b=0}^{B-1} \nabla f_j (\mathbf{y}_{\lfloor t \rfloor, b}^{(j)}) \dd{t}\Big\|^2 \Big]
    \label{eq:thm_2}
\end{alignb}}
where (a) uses Lemma \ref{lemma:ytbj_xtj}, Lemma \ref{lemma:xt0j_xtj}, and Lemma \ref{lemma:dev_local_grads} on the first three terms, respectively.

Considering the third term in the inequality \ref{proof:mainthm-1},
{\allowdisplaybreaks
\begin{alignb}
    &\mathbb{E}_{\cdot|\mathcal{Q}} \Big[ \Big\| \int_{P} \sum_{j\neq i} q_t^{j\rightarrow i} \sum_{b=0}^{B-1} \mathbf{g}_j(\mathbf{y}_{\lfloor t \rfloor, b}^{(j)}) \dd{t} \Big\|^2 \Big] \\
    %&\overset{(a)}{\leq}\mathbb{E}_{\cdot|\mathcal{Q}} \Big[ \Big\| \int_{P} \sum_{j\neq i} q_t^{j\rightarrow i} \sum_{b=0}^{B-1} \mathbf{g}_j(\mathbf{y}_{t, b}^{(j)}) \dd{t}\Big\|^2 \Big] \\
    &=\mathbb{E}_{\cdot|\mathcal{Q}} \Big[ \Big\| \int_{P} \sum_{j\neq i} q_t^{j\rightarrow i} \sum_{b=0}^{B-1} \nabla f_j(\mathbf{y}_{\lfloor t\rfloor, b}^{(j)}) \dd{t}\Big\|^2 \Big] \\
        &\quad+ \mathbb{E}_{\cdot|\mathcal{Q}} \Big[ \Big\| \int_{P} \sum_{j\neq i} q_t^{j\rightarrow i} \sum_{b=0}^{B-1} [\mathbf{g}_j(\mathbf{y}_{\lfloor t\rfloor,b}^{(j)}) - \nabla f_j(\mathbf{y}_{\lfloor t\rfloor, b}^{(j)}) ] \dd{t} \Big\|^2 \Big] \\
    &=\mathbb{E}_{\cdot|\mathcal{Q}} \Big[ \Big\| \int_{P} \sum_{j\neq i} q_t^{j\rightarrow i} \sum_{b=0}^{B-1} \nabla f_j(\mathbf{y}_{\lfloor t\rfloor, b}^{(j)}) \dd{t} \Big\|^2 \Big] \\
        &\quad+  \int_{P} \sum_{j\neq i} (q_t^{j\rightarrow i})^2   \sum_{b=0}^{B-1} \mathbb{E}_{\cdot|\mathcal{Q}} \big[ \|  \mathbf{g}_j(\mathbf{y}_{\lfloor t\rfloor,b}^{(j)}) - \nabla f_j(\mathbf{y}_{\lfloor t\rfloor, b}^{(j)})  \|^2 \big] \dd{t}\\ 
    &\overset{(a)}\leq \mathbb{E}_{\cdot|\mathcal{Q}} \Big[ \Big\| \int_{P} \sum_{j\neq i} q_t^{j\rightarrow i} \sum_{b=0}^{B-1} \nabla f_j(\mathbf{y}_{\lfloor t\rfloor, b}^{(j)}) \dd{t}\Big\|^2 \Big] + B\rho^2\sigma^2\label{eq:thm_3}
\end{alignb}}
where (a) is derived from the definition of $\sigma$ in Assumption \ref{assmp:sg} and $\rho$.%, and (b) is attributed to Lemma \ref{lemma:floor_t_and_t}.

Plugging \ref{eq:thm_2} and \ref{eq:thm_3}, the inequality \ref{proof:mainthm-1} is rephrased as:
{\allowdisplaybreaks
\begin{alignb}
    &\mathbb{E}_{\cdot|\mathcal{Q}}[f_i(\mathbf{x}^{(i)}_{t_0+P})] \\
    &\leq \mathbb{E}_{\cdot|\mathcal{Q}}[f_i(\mathbf{x}^{(i)}_{t_0})] +\gamma \Bigg[ \frac{87BN\zeta^2\Psi}{N-4} +\frac{63BN\sigma^2\Psi}{32} +\frac{81BN^2\zeta^2\Psi}{4} +72B^2L^2N\gamma^2\sigma^2\Psi \\
        &\quad +B\Psi\Big(\frac{21}{32N}+576B^2L^2N\gamma^2 -\frac{1}{2}\Big) \mathbb{E}_{\cdot|\mathcal{Q}} \big[\big\| \nabla f_i(\mathbf{x}_{t_0}^{(i)}) \big\|^2\big]\\
        &\quad -\frac{1}{2B\Psi} \mathbb{E}_{\cdot|\mathcal{Q}} \Big[ \Big\| \int_{P} \sum_{j\neq i} q_t^{j\rightarrow i} \sum_{b=0}^{B-1} \nabla f_j (\mathbf{y}_{\lfloor t \rfloor, b}^{(j)}) \dd{t}\Big\|^2 \Big] \Bigg]\\
        &\quad + \frac{\gamma^2L}{2}\mathbb{E}_{\cdot|\mathcal{Q}} \Big[ \Big\| \int_{P} \sum_{j\neq i} q_t^{j\rightarrow i} \sum_{b=0}^{B-1} \nabla f_j(\mathbf{y}_{\lfloor t\rfloor, b}^{(j)}) \dd{t} \Big\|^2 \Big] +\frac{BL\gamma^2\rho^2\sigma^2}{2} \\
    &\leq \mathbb{E}_{\cdot|\mathcal{Q}}[f_i(\mathbf{x}^{(i)}_{t_0})] + \frac{87BN\gamma\zeta^2\Psi}{N-4} +\frac{63BN\gamma\sigma^2\Psi}{32} +\frac{81BN^2\gamma\zeta^2\Psi}{4} +72B^2L^2N\gamma^3\sigma^2\Psi \\
        &\quad +B\gamma\Psi\Big(\frac{21}{32N}+576B^2L^2N\gamma^2 -\frac{1}{2}\Big) \mathbb{E}_{\cdot|\mathcal{Q}} \big[\big\| \nabla f_i(\mathbf{x}_{t_0}^{(i)}) \big\|^2\big]\\
        &\quad +\Big(\frac{\gamma^2L}{2}-\frac{\gamma}{2B\Psi}\Big) \mathbb{E}_{\cdot|\mathcal{Q}} \Big[ \Big\| \int_{P} \sum_{j\neq i} q_t^{j\rightarrow i} \sum_{b=0}^{B-1} \nabla f_j (\mathbf{y}_{\lfloor t \rfloor, b}^{(j)}) \dd{t}\Big\|^2 \Big] +\frac{BL\gamma^2\rho^2\sigma^2}{2}\\
    &\overset{(a)}{\leq} \mathbb{E}_{\cdot|\mathcal{Q}}[f_i(\mathbf{x}^{(i)}_{t_0})] + \frac{87BN\gamma\zeta^2\Psi}{N-4} +\frac{63BN\gamma\sigma^2\Psi}{32} +\frac{81BN^2\gamma\zeta^2\Psi}{4} +72B^2L^2N\gamma^3\sigma^2\Psi \\
        &\quad +B\gamma\Psi\Big(\frac{21}{32N}+576B^2L^2N\gamma^2 -\frac{1}{2}\Big) \mathbb{E}_{\cdot|\mathcal{Q}} \big[\big\| \nabla f_i(\mathbf{x}_{t_0}^{(i)}) \big\|^2\big] +\frac{BL\gamma^2\rho^2\sigma^2}{2}\\
    %&\overset{(b)}{\leq} \mathbb{E}_{\cdot|\mathcal{Q}}[f_i(\mathbf{x}^{(i)}_{t_0})] +\frac{81BN^2n_P\gamma\zeta^2}{4} + 72B^2L^2Nn_P\gamma^3\sigma^2 +\frac{45BNn_P\gamma\zeta^2}{N-4} \\ &\quad+\frac{21BNn_P\gamma\sigma^2}{1000}  +2Bn_P\gamma\big(36BL^2N\gamma^2+5\big) \mathbb{E}_{\cdot|\mathcal{Q}} [\| \nabla f_i(\mathbf{x}_{t_0}^{(i)})\|^2] +\frac{BL\gamma^2\rho^2\sigma^2}{2} \\
    % simplify the coefficient in front of E[\nabla f_i ...]
    &\overset{(b)}{\leq } \mathbb{E}_{\cdot|\mathcal{Q}}[f_i(\mathbf{x}^{(i)}_{t_0})] + \frac{87BN\gamma\zeta^2\Psi}{N-4} +\frac{63BN\gamma\sigma^2\Psi}{32} +\frac{81BN^2\gamma\zeta^2\Psi}{4} +72B^2L^2N\gamma^3\sigma^2\Psi \\
        &\quad +B\gamma\Psi\Big(\frac{21}{32N}+\frac{9}{N\Psi^2} -\frac{1}{2}\Big) \mathbb{E}_{\cdot|\mathcal{Q}} \big[\big\| \nabla f_i(\mathbf{x}_{t_0}^{(i)}) \big\|^2\big] +\frac{BL\gamma^2\rho^2\sigma^2}{2}, %\\
    %&\leq \mathbb{E}_{\cdot|\mathcal{Q}}[f_i(\mathbf{x}^{(i)}_{t_0})] + \frac{87BN\gamma\zeta^2\Psi}{N-4} +\frac{63BN\gamma\sigma^2\Psi}{32} +\frac{81BN^2\gamma\zeta^2\Psi}{4} +72B^2L^2N\gamma^3\sigma^2\Psi \\
    %    &\quad +BN\gamma\Psi\Big(\frac{21}{32N}+\frac{9}{N\Psi^2} -\frac{1}{2}\Big) \mathbb{E}_{\cdot|\mathcal{Q}} \big[\big\| \nabla f_i(\mathbf{x}_{t_0}^{(i)}) \big\|^2\big] +\frac{BL\gamma^2\rho^2\sigma^2}{2},
    \label{eq:thm_before_arrangement}
\end{alignb}}
where (a) negates the term including $\nabla f_j (\mathbf{y}_{t, b}^{(j)})$ because $\frac{\gamma^2L}{2}-\frac{\gamma}{2B\Psi} < 0$ based on the upper bound of $\gamma$; (b) bounds the coefficient of the term including $\mathbb{E}_{\cdot|\mathcal{Q}} [\| \nabla f_i(\mathbf{x}_{t_0}^{(i)})\|^2]$ to simplify the further analysis.

After rearrangement, we have
\allowdisplaybreaks{\begin{alignb}
    &\mathbb{E}_{\cdot|\mathcal{Q}} [\| \nabla f_i(\mathbf{x}_{t_0}^{(i)})\|^2] \\
    %&\leq \frac{\mathbb{E}_{\cdot|\mathcal{Q}}[f_i(\mathbf{x}^{(i)}_{t_0})] - \mathbb{E}_{\cdot|\mathcal{Q}}[f_i(\mathbf{x}^{(i)}_{t_0+P})]}{B\gamma\Psi\Big(\frac{1}{2} -\frac{21}{32N}-\frac{9}{N\Psi^2}\Big)} + \frac{1}{B\gamma\Psi\Big(\frac{1}{2} -\frac{21}{32N}-\frac{9}{N\Psi^2}\Big)} \Bigg( \frac{87BN\gamma\zeta^2\Psi}{N-4} +\frac{63BN\gamma\sigma^2\Psi}{32} +\frac{81BN^2\gamma\zeta^2\Psi}{4} \\ &\quad+72B^2L^2N\gamma^3\sigma^2\Psi +\frac{BL\gamma^2\rho^2\sigma^2}{2} \Bigg)\\
    &\leq \frac{\mathbb{E}_{\cdot|\mathcal{Q}}[f_i(\mathbf{x}^{(i)}_{t_0})] - \mathbb{E}_{\cdot|\mathcal{Q}}[f_i(\mathbf{x}^{(i)}_{t_0+P})]}{B\gamma\Psi\Big(\frac{1}{2} -\frac{21}{32N}-\frac{9}{N\Psi^2}\Big)} + \frac{1}{\frac{1}{2} -\frac{21}{32N}-\frac{9}{N\Psi^2}} \Bigg( \frac{87N\zeta^2}{N-4} +\frac{63N\sigma^2}{32} +\frac{81N^2\zeta^2}{4} \\
        &\quad+72BL^2N\gamma^2\sigma^2 +\frac{L\gamma\rho^2\sigma^2}{2\Psi} \Bigg)\\
    &\overset{(a)}{\leq} \frac{128\big(\mathbb{E}_{\cdot|\mathcal{Q}}[f_i(\mathbf{x}^{(i)}_{t_0})] - \mathbb{E}_{\cdot|\mathcal{Q}}[f_i(\mathbf{x}^{(i)}_{t_0+P})]\big)}{11B\gamma\Psi} + \frac{128}{11} \Bigg( \frac{87N\zeta^2}{N-4} +\frac{63N\sigma^2}{32} +\frac{81N^2\zeta^2}{4} \\
    &\quad+72BL^2N\gamma^2\sigma^2 +\frac{L\gamma\rho^2\sigma^2}{2\Psi} \Bigg)\\
    &= \frac{128\big(\mathbb{E}_{\cdot|\mathcal{Q}}[f_i(\mathbf{x}^{(i)}_{t_0})] - \mathbb{E}_{\cdot|\mathcal{Q}}[f_i(\mathbf{x}^{(i)}_{t_0+P})]\big)}{11B\gamma\Psi} + \frac{11136N\zeta^2}{11(N-4)} +\frac{252N\sigma^2}{11} +\frac{2592N^2\zeta^2}{11} \\
        &\quad+9216BL^2N\gamma^2\sigma^2 +\frac{64L\gamma\rho^2\sigma^2}{11\Psi}\ ,
    \label{eq:lastpang}
\end{alignb}}
where (a) makes the denominator smaller than the derived upper bound of inequality \ref{eq:thm_before_arrangement} by using $N>4$ and $\Psi\geq3$, resulting in
\[\frac{1}{\frac{1}{2} -\frac{21}{32N}-\frac{9}{N\Psi^2}} \leq \frac{1}{\frac{1}{2}-\frac{21}{32\cdot4}-\frac{9}{4\cdot3^2}} =\frac{128}{11} .\]
%where (a) makes the denominator smaller than the derived upper bound of inequality \ref{eq:thm_before_arrangement}; (b) uses the fact that $n_P\geq1$ and $\gamma\leq\frac{1}{8BL}$ on the fourth and the sixth term, respectively.

Finally, the minimum value of $\mathbb{E}_{\cdot|\mathcal{Q}} [\| \nabla f(\mathbf{x}_t)\|^2]$ over time $t$ can be found as:
\allowdisplaybreaks{\begin{align*}
    &\min_t \mathbb{E}_{\cdot|\mathcal{Q}} [\| \nabla f(\mathbf{x}_t)\|^2] \\
    &= \min_t \mathbb{E}_{\cdot|\mathcal{Q}} \Big[\Big\| \frac{1}{N} \sum_{i=1}^N \nabla f_i(\mathbf{x}_t^{(i)}) \Big\|^2\Big] \\
    &\leq \min_{t_0 \in \{0,P,\cdots,(\lfloor \frac{T}{P} \rfloor-1)P\}} \mathbb{E}_{\cdot|\mathcal{Q}} \Big[\Big\| \frac{1}{N} \sum_{i=1}^N \nabla f_i(\mathbf{x}_{t_0}^{(i)}) \Big\|^2\Big] \\
    &\overset{(a)}{\leq} \min_{t_0 \in \{0,P,\cdots,(\lfloor \frac{T}{P} \rfloor-1)P\}} \frac{1}{N}\cdot \mathbb{E}_{\cdot|\mathcal{Q}} \Big[\sum_{i=1}^N\big\|  \nabla f_i(\mathbf{x}_{t_0}^{(i)}) \big\|^2\Big] \\
    &\leq \frac{1}{N\lfloor \frac{T}{P} \rfloor} \cdot \sum_{t_0=0,P,2P,\cdots,(\lfloor \frac{T}{P} \rfloor-1)P} \mathbb{E}_{\cdot|\mathcal{Q}} \Big[\sum_{i=1}^N\big\|  \nabla f_i(\mathbf{x}_{t_0}^{(i)}) \big\|^2\Big]\\
    &\overset{(b)}{\leq} \frac{1}{N\lfloor \frac{T}{P} \rfloor} \sum_{i=1}^N \Bigg[ \frac{128\big(f_i(\mathbf{x}^{(i)}_0) - f_i^*\big)}{11B\gamma\Psi}\Bigg]  + \frac{11136\zeta^2}{11(N-4)} +\frac{252\sigma^2}{11} +\frac{2592N\zeta^2}{11} \\
        &\quad+9216BL^2\gamma^2\sigma^2 +\frac{64L\gamma\rho^2\sigma^2}{11N\Psi} \\
    &\overset{(c)}{\leq} \frac{128}{11B\gamma\Psi} (f(\mathbf{x}_0)-f^*) +\frac{11136\zeta^2}{11(N-4)} +\frac{252\sigma^2}{11} +\frac{2592N\zeta^2}{11} \\
    &\quad+9216BL^2\gamma^2\sigma^2 +\frac{64L\gamma\rho^2\sigma^2}{11N\Psi} \\
    &=\mathcal{O}\Big(\frac{\mathcal{F}}{B\gamma\Psi} +\frac{\zeta^2}{N-4} +\sigma^2 +N\zeta^2 +BL^2\gamma^2\sigma^2 +\frac{L\gamma\rho^2\sigma^2}{N\Psi}\Big)
\end{align*}}
where (a) is due to Jensen's inequality; the first term of (b) is an implantation of inequality \ref{eq:lastpang} whereas the other terms are independent on $t_0$; (c) takes that $P\leq T$ and the definition of $f(\mathbf{x}_\star)$

%===============

\section{Pseudo algorithm of Draco}\label{appx:psuedo-alg}
\begin{figure}[htbp]
\scalebox{0.99}{
    \includegraphics[width=\textwidth]{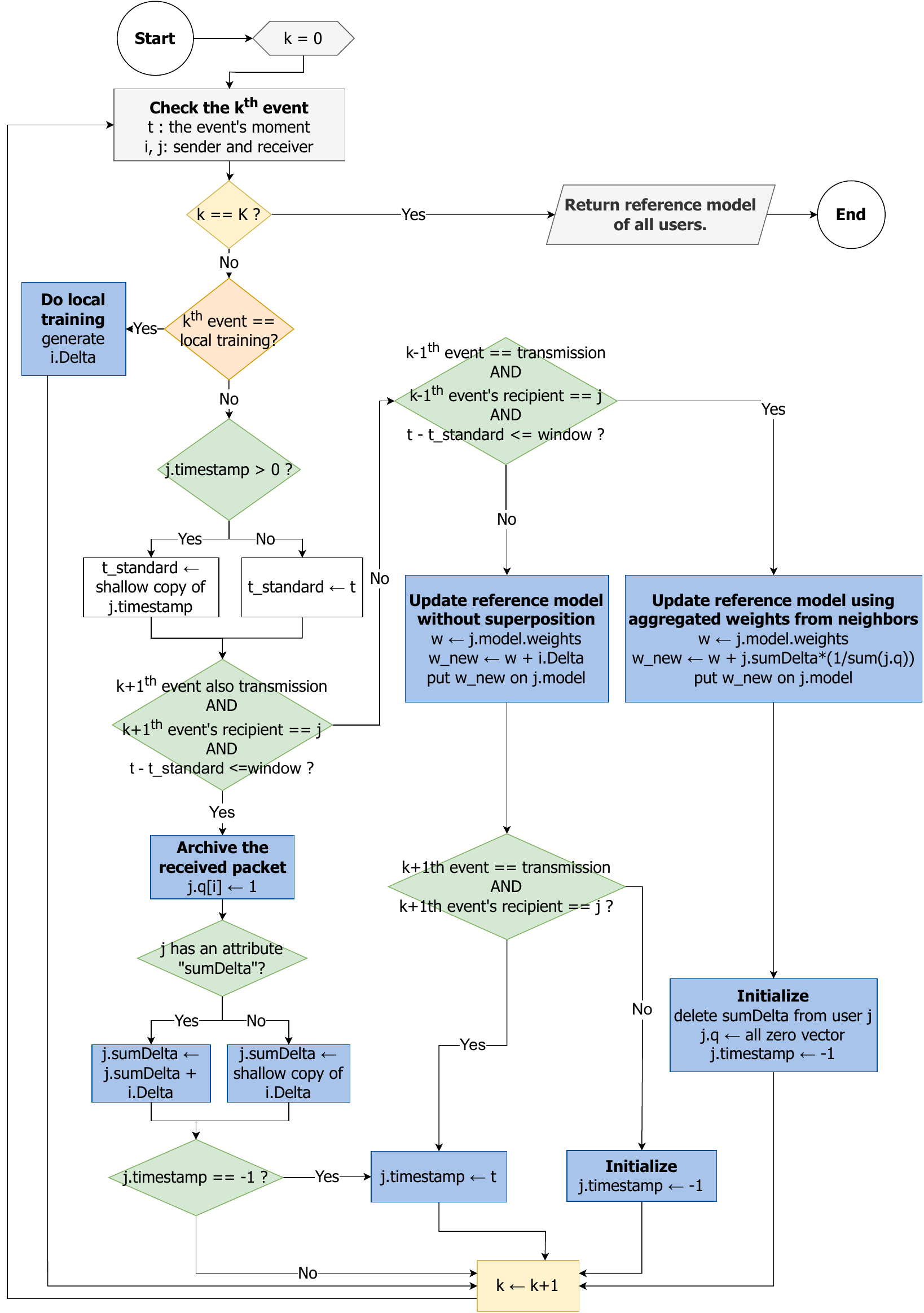}}
    \caption{Flowchart of DRACO after initialization}
    \label{fig:flowchart}
\end{figure}

In this section, we provide a pseudo-algorithm and a flowchart for the intuitive reproduction of source codes. The flowchart in Fig. \ref{fig:flowchart} includes only the transmission/reception procedure of DRACO, corresponding to lines 17-28 (excluding periodic unification parts) of Algorithm \ref{alg:draco-pseudo}.

\begin{figure}[ht]
  \centering
  \scalebox{0.98}{
    \begin{minipage}{\linewidth}
      \begin{algorithm}[H]
        \SetAlgoLined
        \caption{\label{alg:draco-pseudo} Pseudo algorithm of Algorithm \ref{alg:draco-userview}.}
        \BlankLine
        \DontPrintSemicolon
    \KwData{$\gamma, \lambda, \mathbf{x}_0, B, T, P$}
    \KwResult{$\{\mathbf{x}_t : \forall t\}$}
    \Init{}{
        \For{$i=1,\cdots,N$}{
        \tcc{Generate \texttt{ListEvents}$(i)$}
            Generate $t\sim Exp(\lambda_{i})$\;
            Append $[t,i]$ to \texttt{ListEventsGrad}$(i)$\;
            \For{event in \texttt{ListEventsGrad}$(i)$}{
                %$\mathbf{q}^i \sim p(\mathcal{U}\setminus \{i\}, \cdots)$\;
                \For{$j\in\mathcal{N}(i)$}{
                    Generate $t\sim exp(\lambda_{ij})$ or $t\leftarrow$transmission delay\;
                    Append $[t,j]$ to \texttt{ListEventsComm}$(i)$\;
                }
            } 
        $\texttt{ListEvents}(i)\leftarrow\texttt{ListEventsGrad}(i)+\texttt{ListEventsComm}(i)$\;
        }
        \tcc{Generate \texttt{ListEvents} over all clients}
        \For{$i=1,\cdots,N$}{
            Stack \texttt{ListEvents}$(i)$ on \texttt{ListEvents}\;
        }
        Sort \texttt{ListEvents} by $t$ in ascending order.\;
        Add the event indices $k$ in front of each element.\;
    }
    
    $K\leftarrow | \texttt{ListEvents} |$\;
    \For{$k=1,\cdots,K$}{
        $(i,j)\leftarrow \texttt{ListEvents}(k,0),\ \texttt{ListEvents}(k,1)$\;
        \If{$i==j$}{ 
            \For{$b=0,\cdots,B-1$}{
            $\mathbf{y}_{b+1}^{(i)}\leftarrow \mathbf{y}_b^{(i)}-\gamma g_i(\mathbf{y}_b^{(i)})$ \tcp*{local batch training}
            }
            $\Delta_k^{(i)}\leftarrow \mathbf{y}_{k,B}^{(i)} - \mathbf{x}_k^{(i)}$ \;
        }
        \Else{
            %$\mathcal{N}_t(i)\sim P[\delta_{ij}=1|\delta^i=1]$ \tcp*{$k$-th event is a transmission}
            \For{$j\in\mathcal{U}\setminus\{i\}$}{
                \If{\texttt{event\_code}==\textrm{``unification''}}{
                    $\mathbf{x}^{(j)}\leftarrow \Tilde{\mathbf{x}}^{(hub)}$\;
                }
                \Else{
                    $\mathbf{x}^{(j)} \leftarrow \mathbf{x}^{(j)} + q_k^{i\rightarrow j}\Tilde{\Delta}^{(i)}$ \tcp*{aggregation}
                }
            }
        }
    }
    \If{$t\equiv0$ (mod $P$) and $t>0$ and $i$ is the hub}{
        $\mathbf{x}^{(hub)} \leftarrow \mathbf{x}^{(i)}$ \;
        }
    \Return $\mathbf{x}^{(i)}$ for $i\in\mathcal{U}$
      \end{algorithm}
    \end{minipage}
  }
\end{figure} 

\end{document}